\documentclass[journal,final]{IEEEtran}
\usepackage{amsmath,amsfonts}
\usepackage{algorithm}
\usepackage{algpseudocode}
\usepackage{booktabs}
\usepackage{array}
\usepackage[caption=false,font=normalsize,labelfont=sf,textfont=sf]{subfig}
\usepackage{textcomp}
\usepackage{stfloats}
\usepackage{url}
\usepackage{verbatim}
\usepackage{graphicx}
\usepackage{multirow}
\usepackage{cite}
\usepackage{xcolor}
\usepackage[breaklinks]{hyperref}

\usepackage{placeins}
\def\UrlBreaks{\do\/\do-\do@\do.}  
\usepackage{tabularx}
\usepackage{array}
\usepackage{makecell}
\usepackage{orcidlink}

\hypersetup{
    colorlinks=true,
    linkcolor=blue,   
    citecolor=blue,   
    urlcolor=black     
}

\makeatletter
\def\@cite#1#2{%
  {\color{blue}[#1\if@tempswa , #2\fi]}%
}
\makeatother

\begin{document}

\title{A Unified Score Matching Paradigm for Video Anomaly Detection and Anticipation}

\author{Congqi Cao~\orcidlink{0000-0002-0217-9791},~\IEEEmembership{Member,~IEEE,} Zhenhe Liang~\orcidlink{0009-0009-4030-0298}, Hanwen Zhang~\orcidlink{0009-0007-3761-8495}, Yifan Zhao~\orcidlink{0009-0000-0145-2920}, Qinyi Lv~\orcidlink{0000-0002-5359-5701}, \\Lingtong Min~\orcidlink{0000-0003-3970-7823},~\IEEEmembership{Member,~IEEE,} and Yanning Zhang~\orcidlink{0000-0002-2977-8057},~\IEEEmembership{Fellow,~IEEE}
\thanks{This work is supported by National Natural Science Foundation of China (No. 62376217, 62301434, 62576279), the Joint Research Project for Meteorological Capacity Improvement (Grants 24NLTSZ003) and Young Elite Scientists Sponsorship Program by CAST (No. 2023QNRC001).}
\thanks{Congqi Cao, Zhenhe Liang, Hanwen Zhang, Yifan Zhao, and Yanning Zhang are with the National Engineering Laboratory for Integrated Aero-Space-Ground-Ocean Big Data Application Technology, School of Computer Science, Northwestern Polytechnical University, Xi'an 710129, China. Qinyi Lv and Lingtong Min are with the School of Electronics and Information, Northwestern Polytechnical University, Xi'an 710129, China. (e-mail: \url{congqi.cao@nwpu.edu.cn}; \url{zhenheliang@mail.nwpu.edu.cn}; \url{zhwwww@mail.nwpu.edu.cn}; \url{yfzhao@mail.nwpu.edu.cn}; \url{lvqinyi@nwpu.edu.cn}; \url{minlingtong@nwpu.edu.cn}; \url{ynzhang@nwpu.edu.cn}).}
}


\maketitle
\markboth{Journal of \LaTeX\ Class Files,~Vol.~14, No.~8, August~2021}%
{Cao \MakeLowercase{\textit{et al.}}: A Unified Score Matching Paradigm for Video Anomaly Detection and Anticipation}
\pagestyle{headings} 
\begin{abstract}
Video anomaly detection (VAD) is a fundamental and safety-critical task in computer vision. Recent generative approaches detect anomalies from a distributional perspective, but remain limited by local anomaly modes. Meanwhile, video anomaly anticipation (VAA), as a proactive extension beyond post-hoc detection, introduces additional challenges. In particular, the contrastive inference paradigm in VAD, which relies on ground-truth frames, is not applicable to VAA, hindering its development. 
To address these challenges, we propose a unified score-driven framework, termed Uni-DSM, based on denoising score matching (DSM), which models anomaly patterns through likelihood estimation and score functions over the learned data distribution. 
Within this unified framework, we adopt a shared noise-conditioned score transformer backbone with scene-dependent embeddings and motion-aware weighting for distribution-level modeling. Instead of introducing separate architectures, Uni-DSM unifies VAD and VAA through different inference and supervision paradigms built upon the same score-based formulation. 
For VAD, we instantiate an autoregressive denoising score matching (ADSM) mechanism, which progressively accumulates anomalous evidence via autoregressive denoising, enabling enhanced perception of local modes beyond visual cues. For VAA, we extend the same architecture by incorporating a lightweight auxiliary decoder and a novel self-distilled denoising score matching (SDSM) mechanism. By constructing supervision from output discrepancies instead of relying on unavailable future ground truth, our method achieves efficient training suitable or early anomaly anticipation. 
Extensive experiments on multiple benchmark datasets demonstrate state-of-the-art performance in both VAD and VAA while maintaining high efficiency, establishing a unified and scalable pipeline from anomaly detection to anticipation.
\end{abstract}


\section{Introduction}
\IEEEPARstart{V}{ideo} anomaly detection (VAD) and video anomaly anticipation (VAA) are core tasks in intelligent surveillance. Compared to VAD, which detects anomalies after occurrence, VAA forecasts near-future risks to enable proactive response, which is particularly critical in high-risk scenarios.

Due to the rarity and unpredictability of anomalous events, existing approaches generally adopt a semi-supervised paradigm trained solely on normal data~\cite{chandola2009anomaly}, \cite{adam2008robust}, \cite{saligrama2010video}, \cite{ramachandra2020survey}. In this setting, reconstruction-based methods~\cite{gong2019memorizing}, \cite{nguyen2019anomaly}, \cite{zaheer2020old}, \cite{hasan2016learning}, \cite{cao2024context}, \cite{wang2022video}, \cite{ionescu2019object} 
and prediction-based methods~\cite{cao2023new}, \cite{cao2024scene}, \cite{liu2018future}, \cite{cai2021appearance}, \cite{liu2021hybrid}, \cite{li2021variational}, \cite{rodrigues2020multi}, \cite{georgescu2021anomaly}, \cite{park2020learning}, \cite{ristea2024self}, \cite{lv2021learning} have emerged as two dominant paradigms. Their core assumption is that anomalous samples are more difficult to accurately reconstruct and predict than normal ones~\cite{hasan2016learning, liu2018future}. However, these methods primarily rely on low-level visual contrastive errors, which makes them prone to misclassifying samples with significant appearance variations but normal semantics as anomalies.

\begin{figure}
  \centering
  \includegraphics[width=0.98\columnwidth]{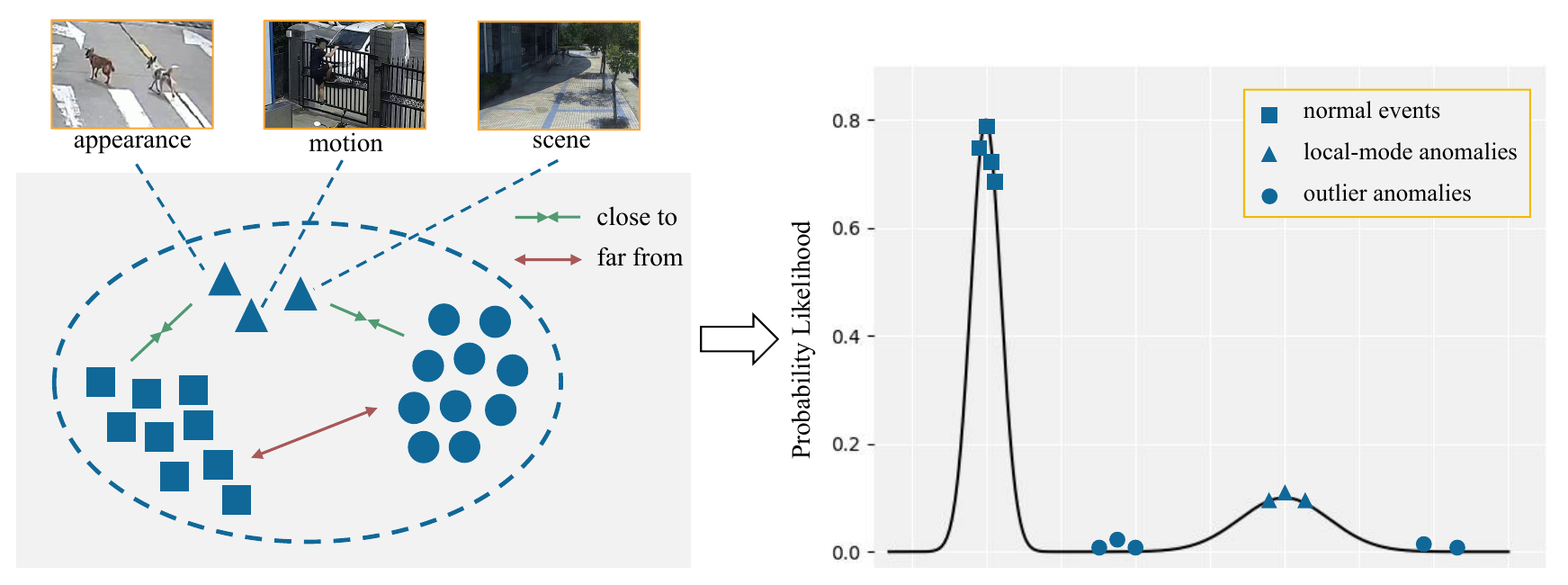}
  \caption{An illustration of the local modes of anomalies. We explore two regions in which the anomalies are residing. The likelihood-based method is highly sensitive to regions of anomalies with inherently low probability likelihood while blind to anomalies located in local modes near the learned distribution.}
  \label{fig:fig1}
\end{figure}

\IEEEpubidadjcol

In recent years, advances in generative models~\cite{ho2020denoising}, \cite{song2019generative}, \cite{song2020score}, \cite{kingma2021variational}, \cite{dhariwal2021diffusion}, \cite{saharia2022image} have offered new perspectives for video anomaly modeling: anomalous events inherently lie in low-probability density regions~\cite{zong2018deep} of the normal data distribution and can be characterized via likelihoods or their gradients. Although prior studies have explored leveraging likelihood information for auxiliary visual pretraining~\cite{flaborea2023multimodal, yan2023feature} or density estimation~\cite{hirschorn2023normalizing}, \cite{mahmood2020multiscale}, \cite{micorek2024mulde}, they have not fully exploited its potential for direct anomaly modeling. However, directly relying on log-likelihoods or their gradients (i.e., the Stein score~\cite{liu2016kernelized}) for anomaly detection remains challenging. As shown in Fig.~\ref{fig:fig1}, the Stein score tends to vanish at extrema and saddle points of the distribution, making anomalies within local modes difficult to distinguish effectively~\cite{mahmood2020multiscale}. 
More fundamentally, we attribute this limitation to several gaps between likelihood-based measures and visual perception, including a scene gap that is insensitive to semantic context~\cite{cao2023new, cao2024scene}, a motion gap due to insufficient modeling of dynamic patterns, and an appearance gap stemming from limited capability to capture fine-grained visual variations.

\begin{figure*}[t]  
  \centering
  \includegraphics[width=\textwidth]{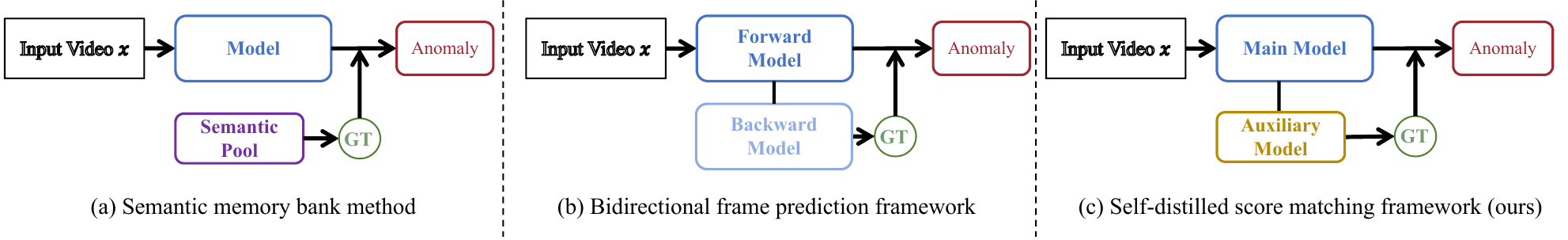}  
  \caption{An illustration of three video anomaly anticipation frameworks, where GT denotes ground truth. Compared with semantic memory bank and bidirectional frame prediction frameworks, our self-distilled DSM (SDSM) within Uni-DSM adopts a lightweight unidirectional main-auxiliary architecture, enabling efficient video anomaly anticipation with lower GPU memory consumption and faster inference speed.}
  \label{fig:fig8}
\end{figure*}

Meanwhile, as a proactive anticipation task, VAA poses even greater challenges.
The primary difficulty lies in the lack of future ground truth, which renders contrastive-objective-based VAD inference paradigms unsuitable. 
As shown in Fig.~\ref{fig:fig8}, prior work addresses this using semantic memory banks~\cite{leng2022anomaly} or forward-backward prediction frameworks~\cite{cao2023new,cao2024context}. 
While bidirectional architectures better capture spatiotemporal context and achieve strong generalization, surpassing semantic-based methods, they significantly increase computational complexity. This limits their applicability in real-world deployment.

\begin{figure*}[t]  
  \centering
  \includegraphics[width=\textwidth]{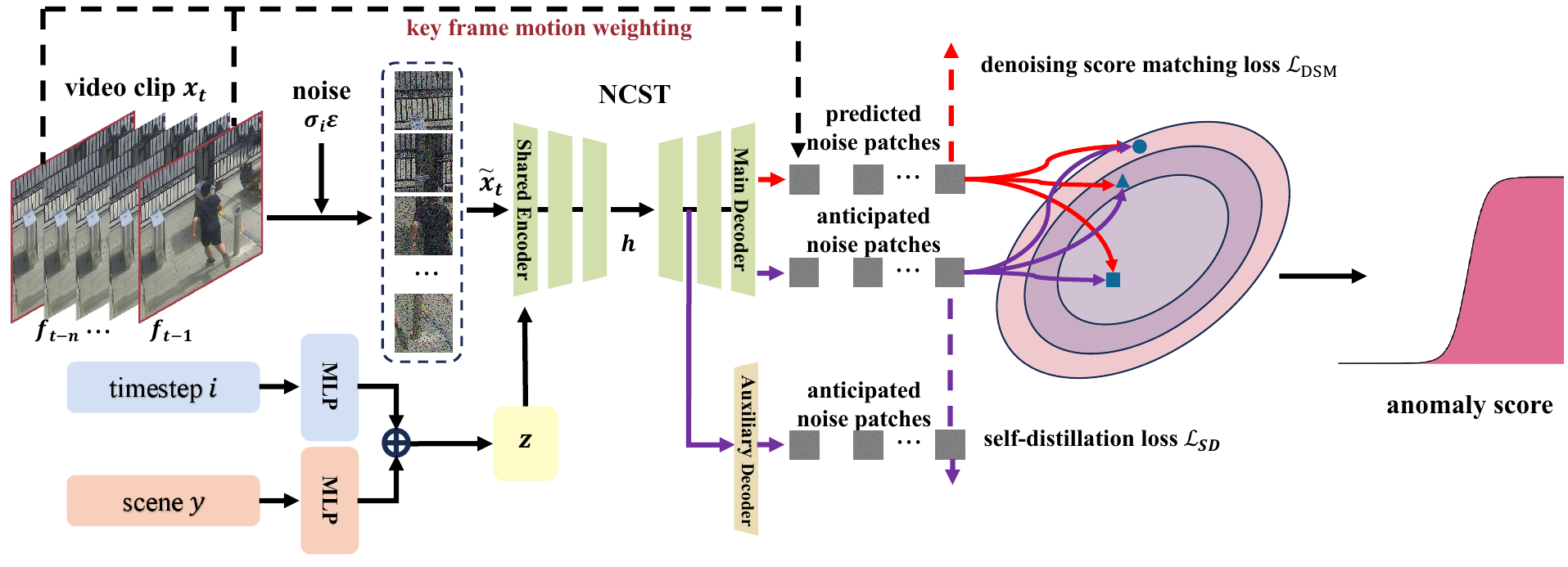}  
  \caption{An overview of the Uni-DSM framework, a unified score-driven architecture based on denoising score matching (DSM). The red flow represents autoregressive denoising score matching (ADSM) for video anomaly detection, while the purple flow denotes self-distilled score matching (SDSM) for video anomaly anticipation.}
  \label{fig:fig3}
\end{figure*}

To address these challenges, as shown in Fig.~\ref{fig:fig3}, we propose a unified score-driven framework, termed Uni-DSM, based on denoising score matching (DSM), which models video anomaly detection and anticipation from a distributional perspective. In this framework, anomalies are characterized as low-likelihood regions under the learned normal distribution, and score functions are leveraged to quantify such distributional deviations. Unlike conventional methods relying on pixel-level contrastive errors, our approach directly operates on the intrinsic data distribution. Building upon this formulation, we formulate a unified score matching paradigm that supports both detection and anticipation through task-aligned inference and supervision mechanisms. 
Within the Uni-DSM framework, we adopt a shared noise-conditioned score transformer (NCST) as the backbone to approximate the Stein score of noisy data under multiple noise levels. The NCST embeds denoising score matching into a diffusion transformer, enabling effective modeling of complex spatiotemporal dependencies. On top of this shared formulation, we further introduce a scene-dependent and motion-aware score by incorporating scene-conditioned embeddings and motion weights derived from key-frame discrepancies, which strengthens temporal consistency and improves robustness under varying contexts.

For the VAD task, we instantiate an autoregressive denoising score matching (ADSM) mechanism. Specifically, we develop an autoregressive denoising strategy during inference that progressively injects intensifying noise and accumulates anomalous context across steps. Meanwhile, discrepancies between denoised outputs and original inputs are integrated to enhance appearance perception and suppress local modes, leading to more reliable anomaly discrimination.

For the VAA task, we extend the same score-driven formulation and propose a self-distilled denoising score matching (SDSM) mechanism. Instead of relying on unavailable future ground-truth frames, SDSM constructs supervision from output discrepancies under a lightweight main-auxiliary architecture with unidirectional temporal inference. This design maintains architectural consistency with ADSM while introducing a self-distillation paradigm tailored for anticipation, enabling efficient training and practical predictive anomaly warning with reduced computational overhead.

In summary, we present the first unified framework that introduces denoising score matching into both video anomaly detection and anticipation under a consistent score-driven formulation. The proposed Uni-DSM framework integrates a shared noise-conditioned score transformer, a scene-dependent and motion-aware score function, and two task-specific mechanisms (e.g., autoregressive DSM for detection and self-distilled DSM for anticipation). Extensive experimental results demonstrate that the proposed methods consistently outperform state-of-the-art approaches on multiple benchmark datasets, including Avenue~\cite{lu2013abnormal}, ShanghaiTech~\cite{luo2017revisit}, and NWPU Campus~\cite{cao2023new}. Our contributions are as follows:
\begin{itemize}

\item A unified score-driven framework, Uni-DSM, based on denoising score matching (DSM), representing the first formulation that models anomalies as low-likelihood patterns under the learned normal distribution for both video anomaly detection and anticipation.

\item A shared score-based modeling strategy with a noise-conditioned score transformer, integrating scene-dependent embeddings and motion-aware weighting to mitigate scene, motion, and appearance gaps and align likelihood estimation with structured visual dynamics.

\item An autoregressive DSM (ADSM) mechanism for VAD, enhancing sensitivity to local anomalies by progressively injecting Gaussian noise and aggregating discrepancies between denoised and original data via the score function.

\item A self-distilled DSM (SDSM) mechanism for VAA, extending the architecture with a lightweight auxiliary decoder to enable ground-truth-free anticipation via self-distillation and unidirectional temporal inference.

\item Extensive experiments on multiple benchmarks demonstrate state-of-the-art performance in both detection and anticipation, confirming the effectiveness and generalization of the unified framework.

\end{itemize}

This paper is an extension of our previous work~\cite{zhang2025autoregressive}. We
extend the preliminary version in the following aspects: (1) We generalize the framework from video anomaly detection to video anomaly anticipation under the same unified denoising score matching formulation, establishing a coherent pipeline for both tasks. (2) We introduce a self-distilled DSM (SDSM) mechanism to address the absence of future ground-truth frames by leveraging output discrepancy. (3) We design a lightweight main-auxiliary architecture with unidirectional temporal inference, which replaces bidirectional paradigms and significantly reduces model complexity. (4) We conduct more comprehensive experiments with thorough analyses, demonstrating consistent improvements in both performance and efficiency, especially in predictive anomaly warning scenarios.

\section{RELATED WORK}
\subsection{Video Anomaly Detection Methods}
Mainstream video anomaly detection (VAD) methods can be broadly categorized into semi-supervised~\cite{sultani2018real}, weakly supervised~\cite{zhong2019graph, zhu2019motion}, and fully supervised~\cite{zhou2016spatial} approaches, depending on the availability of anomaly data and labels during training. This paper focuses on semi-supervised methods, which primarily fall into three paradigms including distance-based methods~\cite{smeureanu2017deep}, \cite{lu2022learnable}, \cite{chang2020clustering}, \cite{wu2019deep}, \cite{xu2015learning}, 
probability-based methods~\cite{cao2024scene}, \cite{li2021variational}, \cite{rodrigues2020multi},~\cite{li2013anomaly}, \cite{hinami2017joint}, and reconstruction-based methods~\cite{gong2019memorizing}, \cite{nguyen2019anomaly}, \cite{zaheer2020old}, \cite{hasan2016learning}, \cite{cao2024context}, \cite{wang2022video}, \cite{ionescu2019object}, \cite{cao2023new}, \cite{liu2018future}, \cite{cai2021appearance}, \cite{liu2021hybrid}, \cite{georgescu2021anomaly}, \cite{park2020learning}, \cite{ristea2024self}, \cite{lv2021learning}, \cite{fang2020anomaly}, \cite{luo2021future}, \cite{song2019learning}, \cite{yu2020cloze}. Reconstruction-based methods include both spatial frame reconstruction and temporal feature prediction. These methods train deep networks via pretext tasks such as reconstructing frames with autoencoders~\cite{gong2019memorizing}, \cite{nguyen2019anomaly}, \cite{zaheer2020old}, \cite{hasan2016learning}, \cite{ionescu2019object}, 
predicting future frames or features~\cite{cao2023new}, \cite{cao2024scene}, \cite{liu2018future}, \cite{cai2021appearance}, \cite{liu2021hybrid}, \cite{li2021variational}, \cite{rodrigues2020multi}, \cite{park2020learning}, \cite{lv2021learning}, \cite{fang2020anomaly}, \cite{luo2021future}, \cite{song2019learning}, 
learning dictionaries~\cite{lu2022learnable}, \cite{chang2020clustering}, \cite{wu2019deep}, \cite{jiang2024effectiveness}, 
and solving masked jigsaw puzzles~\cite{wang2022video}, \cite{georgescu2021anomaly}, \cite{ristea2024self}. 
The strong mode coverage capability of generative models~\cite{flaborea2023multimodal}, \cite{yan2023feature},~\cite{hirschorn2023normalizing}, \cite{micorek2024mulde} has further inspired a novel class of likelihood-based VAD methods. These approaches build generative models to learn the distribution of normal data, either to enhance visual pretraining objectives~\cite{flaborea2023multimodal, yan2023feature} or to assist density estimation models~\cite{hirschorn2023normalizing, micorek2024mulde}. However, despite their effectiveness, directly adopting likelihood-based metrics still suffers from the issue of invisible anomalies residing in local modes, making them difficult to distinguish.

In recent years, diffusion models~\cite{ho2020denoising} with Langevin-based score matching~\cite{song2019generative, song2020score} have shown strong performance. The score function is defined as the gradient of log density with respect to the input and can be viewed as a vector field pointing to higher data density, making it suitable for anomaly detection~\cite{cao2024scene},~\cite{rodrigues2020multi},~\cite{micorek2024mulde}. 
Cao \textit{et al}.~\cite{cao2024scene} build a scene conditioned latent autoencoder for anomaly detection and anticipation. Rodrigues \textit{et al}.~\cite{rodrigues2020multi} approximate scores at multiple noise levels to capture multi-scale gradients. Micorek \textit{et al}.~\cite{micorek2024mulde} show that using video features improves log density estimation but reduces nonlinearity. However, these methods are limited by the local modes of the score function, and relying solely on likelihood or score magnitude is insufficient for robust anomaly discrimination. More importantly, they lack explicit modeling of structured visual dynamics, including scene context and motion patterns, which leads to a misalignment between distribution-level estimation and visual perception.

To address these challenges, we propose an autoregressive denoising score matching (ADSM) mechanism within Uni-DSM. It models temporal dependencies via autoregressive denoising with accumulated abnormal context, while jointly leveraging visual appearance and likelihood for complementary enhancement. We adopt a noise-conditioned score transformer (NCST) as the backbone. Compared with standard DSM, which is insensitive to local-mode anomalies, Uni-DSM introduces scene-dependent and motion-aware constraints into score estimation, producing a more robust anomaly indicator that better separates normal and anomalous events.
\subsection{Video Anomaly Anticipation Methods}
Compared to the well-established video anomaly detection task, video anomaly anticipation (VAA) is significantly more challenging, as it requires forecasting events that have not occurred, and remains at an early stage. As shown in Fig.~\ref{fig:fig8}, existing approaches primarily focus on how to construct a reference for predictive inference in the absence of future ground truth, leading to two main technical paradigms.

Semantic memory bank methods replace missing ground-truth references with dynamically updated feature banks. For instance, EPAP-Net~\cite{leng2022anomaly} learns the semantic distribution of future frames from normal videos and maintains a semantic pool using a fill-and-update strategy, while employing a channel-selective shifting encoder to capture motion dynamics. Although this approach avoids the complexity of pixel-level anticipation, it suffers from weak correlation between stored semantic features and evolving anomaly patterns, and its reliance on a fixed semantic representation limits generalization across diverse scenarios.

Bidirectional frame prediction frameworks adopt an end-to-end data-driven strategy, achieving strong spatiotemporal modeling and outperforming semantic memory bank methods on standard benchmarks. A representative example is FBAE~\cite{cao2023new}, which jointly trains a forward network for future prediction and a backward network for past reconstruction. During inference, the backward network takes forward-predicted frames as input, and the resulting forward-backward reconstruction error serves as the anomaly signal. Despite their effectiveness, such designs introduce additional computational overhead during both training and inference due to the involvement of the backward network. Moreover, the forward-backward interaction and reliance on reconstruction accuracy may pose challenges for maintaining stable performance in long-term anticipation.

To further improve efficiency and better accommodate predictive anomaly anticipation in real-world scenarios, we extend Uni-DSM to video anomaly anticipation by introducing a lightweight main-auxiliary decoder and a self-distilled denoising score matching (SDSM) mechanism. This design maintains architectural consistency with ADSM while enabling ground-truth-free supervision for efficient and accurate predictive anomaly anticipation.

\section{Task Definition}
\begin{figure}[t]
  \centering
  \includegraphics[width=0.98\columnwidth]{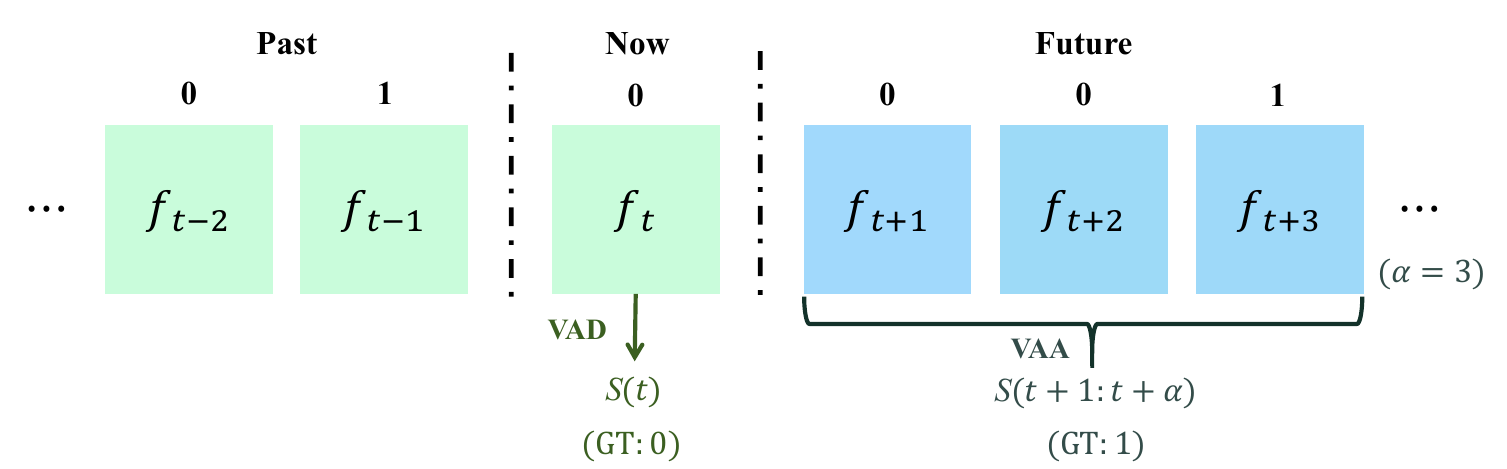}
  \caption{Schematic illustration of the video anomaly detection (VAD) and video anomaly anticipation (VAA) tasks.}
  \label{fig:fig2}
\end{figure}

Video anomaly detection (VAD) aims to determine whether an anomalous event is occurring at the current time, while video anomaly anticipation (VAA) predicts whether an anomaly will occur within a future time window. Compared to VAD, VAA enables proactive early warning without requiring precise prediction of the exact anomaly onset. Fig.~\ref{fig:fig2} illustrates both tasks, where $f_t$ denotes the frame at time $t$, $S(\cdot)$ represents the anomaly score, and $\alpha$ is the anticipation horizon.

\subsection{Video Anomaly Detection}
The VAD task focuses on the immediate judgment of the current state. Given a historical observation sequence $x_{in} = \{f_{t-n}, \dots, f_{t-1}\}$, VAD aims to estimate the anomaly score $S(t)$ for the current frame $f_t$. Let $\mathcal{G}_0 = \{g_t\}_{t=1}^T$ denote the frame-level ground-truth labels, where $g_t \in \{0,1\}$. A normal frame corresponds to a low anomaly score, while an anomalous frame yields a high score. Due to its reactive nature, VAD can only raise alarms after anomalies have occurred.

\subsection{Video Anomaly Anticipation}
Given the same input $x_{in}$, VAA predicts whether an anomaly will occur within the future window $[t+1, t+\alpha]$. The model outputs a score $S(t+1:t+\alpha)$ indicating the probability of at least one anomaly within this interval. Accordingly, the anticipation label sequence is defined as:
\begin{equation}
    \mathcal{G}_\alpha = \left\{ \max\left(\{g_{t+i}\}_{i=1}^\alpha\right) \right\}_{t=1}^{T-\alpha}.
\end{equation}

Unlike VAD, which evaluates the current state, VAA focuses on future risk, leading to different optimization objectives under the same input. Moreover, the absence of future ground-truth frames makes conventional contrastive or prediction-based paradigms inapplicable, motivating the need for alternative supervision strategies.

\section{A Unified Score-Driven Framework Based on Denoising Score Matching}
\subsection{Framework Overview}
As illustrated in Fig.~\ref{fig:fig3}, we propose Uni-DSM, a unified probability-driven framework based on denoising score matching (DSM). Built upon a shared noise-conditioned score transformer (NCST), the framework models anomaly patterns through likelihood estimation and score functions over the learned data distribution. To bridge the gaps across scene, motion, and appearance, we incorporate a scene-dependent and motion-aware score function that embeds scene conditions and assigns motion weights based on key-frame discrepancies. Uni-DSM further includes two task-specific variants under this unified formulation. The autoregressive DSM (ADSM) targets detection and addresses anomalies residing in local modes through autoregressive noise injection. The self-distilled DSM (SDSM) targets anticipation and enables ground-truth-free supervision with a lightweight auxiliary branch. Together, this design forms a coherent pipeline that supports both detection and anticipation within a single architecture.

\subsection{Denoising Score Matching}
\label{sec:dsm}
Let $p(x)$ denote the probability density of normal video data. Its Stein score is defined as the gradient of the log-density: $\nabla_{x} \log p(x)$. Following \cite{vincent2011connection}, we inject i.i.d. Gaussian noise into the clean data $x$ to obtain perturbed samples $\tilde{x}$, and train a score network $s_\theta(\tilde{x})$ using score matching~\cite{hyvarinen2005estimation}. Through this process, we aim to approximate the score function of the corrupted data distribution $q_\sigma(\tilde{x}) = \int q_\sigma(\tilde{x}|x) p(x) \, dx$, where $\sigma$ controls the noise magnitude. It has been theoretically established that this training objective is equivalent to matching the score of a non-parametric Parzen density estimator constructed from the data:
\begin{equation}
    \frac{1}{2} \mathbb{E}_{q_\sigma(\tilde{x}|x) p(x)} \left[ \left\| s_\theta(\tilde{x}) - \nabla_{\tilde{x}} \log q_\sigma(\tilde{x}|x) \right\|^2 \right]
\end{equation}

This approximation can be further extended to generative models~\cite{song2019generative, song2020score}, which progressively perturb data with increasing Gaussian noise levels and train a noise-conditioned score network $s_\theta(x, \sigma)$ to jointly estimate the score functions for all noise levels (i.e., for any $\sigma \in \{\sigma_i\}_{i=1}^L$, $s_\theta(x, \sigma) \approx \nabla_{x} \log q_\sigma(x)$). If the noise distribution is chosen as $\mathcal{N}(\tilde{x}|x, \sigma^2 I)$, the score function simplifies to 
\begin{equation}
\nabla_{\tilde{x}} \log q_\sigma(\tilde{x}|x) = -\frac{\tilde{x} - x}{\sigma^2}
\end{equation}
which can be interpreted as a vector field pointing toward the denoising direction~\cite{vincent2011connection, song2019generative}. Consequently, the training objective becomes:
\begin{equation}
\begin{split}
\frac{1}{L} \sum_{i=1}^{L} \lambda(\sigma_i) \cdot \frac{1}{2} \,
\mathbb{E}_{q_{\sigma_i}(\tilde{x} \mid x) \, p(x)}
\left\| s_\theta(\tilde{x}, \sigma_i) + \frac{\tilde{x} - x}{\sigma_i^2} \right\|^2
\end{split}
\end{equation}
where $\lambda(\sigma_i) > 0$ is a weighting function. Following~\cite{song2019generative}, we set $\lambda(\sigma_i) = \sigma_i^2$.

\begin{figure}[t] 
  \centering
  \includegraphics[width=0.85\columnwidth, height=0.35\textheight, keepaspectratio]{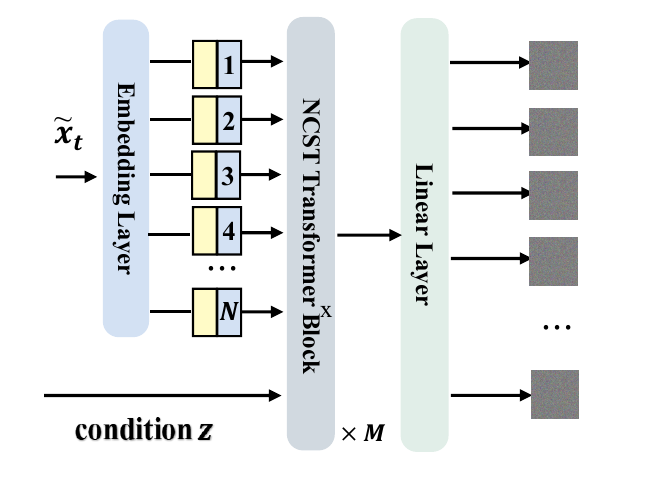}
  \caption{Overview of the proposed noise-conditional score transformer (NCST) architecture used in Fig.~\ref{fig:fig3}.}
  \label{fig:ncst}
\end{figure}

\subsection{Noise-Conditional Score Transformer}
\label{sec:ncst}
To implement denoising score matching for both video anomaly detection and anticipation, we require a model that not only captures the complex spatiotemporal dynamics of video but also conditions on auxiliary variables such as noise level. Attention-based architectures~\cite{vaswani2017attention} provide an intuitive solution for modeling long-range dependencies in video sequences. Moreover, the diffusion transformer (DiT)~\cite{peebles2023scalable} has demonstrated the feasibility of integrating denoising into attention mechanisms. Motivated by this, we propose a novel noise-conditional score transformer (NCST), illustrated in Fig.~\ref{fig:ncst}, which is trained by adapting the DSM objective to a block-wise formulation.

Given an input video sequence $x_t = \{f_{t-n}, \dots, f_{t-1}\}$ and a geometric noise schedule $\{\sigma_i\}_{i=1}^L$ satisfying $\sigma_L/\sigma_{L-1} = \cdots = \sigma_2/\sigma_1 > 1$, we perturb $x_t$ with Gaussian noise $\boldsymbol{\epsilon} \sim \mathcal{N}(0, \sigma_i^2 I)$ at level $\sigma_i$ to obtain $\tilde{x}_t$, where $q_{\sigma_i}(\tilde{x}|x) = \mathcal{N}(\tilde{x}|x, \sigma_i^2 I)$. The $n$ perturbed frames are then uniformly partitioned into $N$ non-overlapping image patches of size $d \times d \times c$, yielding token sequences $\{\tilde{P}_j\}_{j=1}^{N} \in \mathbb{R}^{d \times d \times c}$ and $\{P_j\}_{j=1}^{N} \in \mathbb{R}^{d \times d \times c}$ corresponding to $\tilde{x}_t$ and $x_t$, respectively. Here, $c$ denotes the channel dimension and $d$ is a predefined hyperparameter. For an input frame of spatial resolution $H \times W$, the total number of patches is $N = n \cdot (H/d) \cdot (W/d)$.

Our noise-conditional score transformer $s_\theta(P, \sigma)$ aims to estimate the score for each noisy patch via the block-wise DSM objective:
\begin{equation}
\begin{aligned}
\mathcal{L}_{\text{NCST}} ={}& \frac{1}{L} \sum_{i=1}^{L} \frac{\sigma_i^2}{2} 
\sum_{j=1}^{N} \mathbb{E}_{q_{\sigma_i}(\tilde{P}_j \mid x) p(x)} \\
&\hspace{5em} \left\| s_\theta(\tilde{P}_j, \sigma_i) + \frac{\tilde{P}_j - P_j}{\sigma_i^2} \right\|^2
\end{aligned}
\end{equation}

\begin{figure}[t] 
  \centering
  \includegraphics[width=0.85\columnwidth, height=0.35\textheight, keepaspectratio]{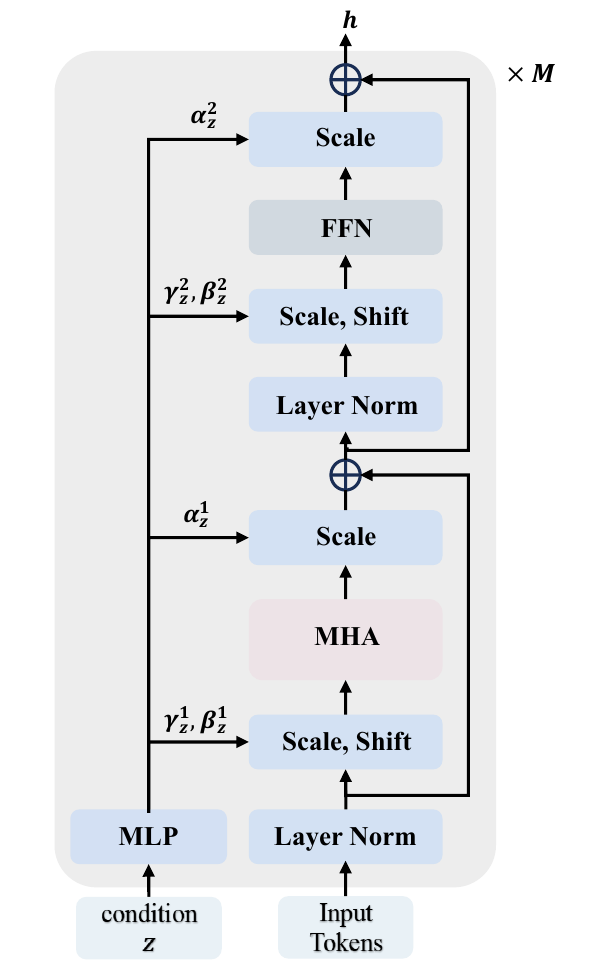}
  \caption{Detailed architecture of the NCST block utilized within the noise-conditional score transformer. MLP, FFN, and MHA stand for the multi-layer perception layer, the feed-forward neural network, and the multi-head attention layer, respectively.}
  \label{fig:fig4}
\end{figure}

The NCST is instantiated within a diffusion-transformer (DiT) framework~\cite{peebles2023scalable, ma2024latte}, which has recently demonstrated strong scalability and representation capacity in generative modeling. We condition the model on the diffusion time step $i$ (which is mapped to $\sigma_i$ via an MLP embedding) to control the noise level. Leveraging the attention-based architecture, continuous supervision across multi-scale block-wise denoising enables the learning of a stable vector field, thereby enhancing the discriminability of anomalies.

\subsection{Scene-Conditioned Embedding}
Scene-dependent anomalies~\cite{ramachandra2020survey} have attracted increasing attention as a crucial characteristic of real-world video anomalies. Since the normality or abnormality of an event is inherently contingent on its specific environment, incorporating scene-aware context significantly improves detection performance~\cite{cao2023new}, \cite{cao2024scene}, \cite{sun2023hierarchical}.

For each video sequence, we assign a scene class label $y$ based on the scene (i.e., the camera ID) to which it belongs. We then combine the scene label $y$ with the diffusion time step $i$ to form the overall condition $z$. 
To incorporate this condition into the transformer backbone, we apply the scalable adaptive layer normalization (S-AdaLN) mechanism~\cite{ma2024latte} to embed and transmit the overall condition $z$. This approach is demonstrated to propagate the condition more effectively throughout the model.

As shown in Fig.~\ref{fig:fig4}, once the conditioning vector $z$ is obtained, it is fed into an MLP to produce a series of scaling and shifting coefficients $\gamma_z^1, \beta_z^1, \alpha_z^1, \gamma_z^2, \beta_z^2$, and $\alpha_z^2$. These coefficients are applied to regulate the hidden features inside each transformer block through a two-step adaptive modulation process:
\begin{subequations}
\begin{align}
    h' &= h + \alpha_z^1 \cdot \text{MHA}\!\left( \gamma_z^1 \cdot \text{LayerNorm}(h) + \beta_z^1 \right) \\
    h  &= h' + \alpha_z^2 \cdot \text{FFN}\!\left( \gamma_z^2 \cdot \text{LayerNorm}(h') + \beta_z^2 \right) 
\end{align}
\end{subequations}
where $h$ and $h'$ are hidden outputs within the transformer blocks. MHA and FFN denote the multi-head attention layer~\cite{vaswani2017attention} and the feed-forward neural network, respectively. By performing block-level adaptive modulation, this conditioning strategy transmits the condition information to each transformer block, contributing to superior performance and faster model convergence.

\subsection{Key-Frame Motion Weighting}
\label{subsec:motion_weighting}
Unlike the Fisher score $\nabla_\theta \log p_\theta(x)$ in classical statistics, which depends on the model parameters $\theta$, the score function in our framework is defined with respect to the input sequence $x_t$. This data-dependent formulation allows the score to be adaptively adjusted according to the characteristics of the observed sequence.
In practical surveillance scenarios, the dynamic foreground and static background of video frames are not balanced~\cite{ramachandra2020survey}. As a result, directly estimated scores tend to be biased toward background statistics. To alleviate this issue, we explicitly inject motion information into the denoising score matching process.

Following insights from video coding principles~\cite{liu2010key}, we estimate motion by computing the absolute difference between the first and last frames of the input sequence $x_t = \{f_{t-n}, \dots, f_{t-1}\}$. This yields a motion representation:
\begin{equation}
g_t = |f_{t-1} - f_{t-n}|
\label{eq:gradient_diff}
\end{equation}
The resulting motion map is divided into $N$ non-overlapping patches $\{G_j\}_{j=1}^N \in \mathbb{R}^{d \times d \times c}$. For each patch, a motion saliency score $v_j$ is computed by first applying a spatial maximum within each channel and then averaging across channels:
\begin{equation}
v_j = \frac{1}{c} \sum_{l=1}^{c} \max\left(G_{j,l}\right)
\label{eq:channel_pooling}
\end{equation}
where $G_{j,l} \in \mathbb{R}^{d \times d}$ denotes the $l$-th channel of the $j$-th patch. 
These saliency values are subsequently normalized to produce patch-level weights:
\begin{equation}
\omega_j = \frac{v_j}{\sum_{k=1}^{N} v_k}
\label{eq:omega_weight}
\end{equation}
The weighted denoising score matching objective is then expressed as:
\begin{equation}
\begin{aligned}
\mathcal{L}_{\text{DSM}} ={}& \frac{1}{L} \sum_{i=1}^{L} \frac{\sigma_i^2}{2} \sum_{j=1}^{N} \omega_j 
\mathbb{E}_{q_{\sigma_i}(\tilde{P}_j \mid x)\,p(x)} \\
&\hspace{6em} \left\| s_\theta(\tilde{P}_j, \sigma_i) + \frac{\tilde{P}_j - P_j}{\sigma_i^2} \right\|^2
\end{aligned}
\end{equation}

Through this key frame motion weighting mechanism, the denoising score matching procedure assigns greater emphasis to dynamic foreground and reduces interference from static background areas, thereby encapsulating more motion information.

\begin{figure}
  \centering
  \includegraphics[width=0.98\columnwidth]{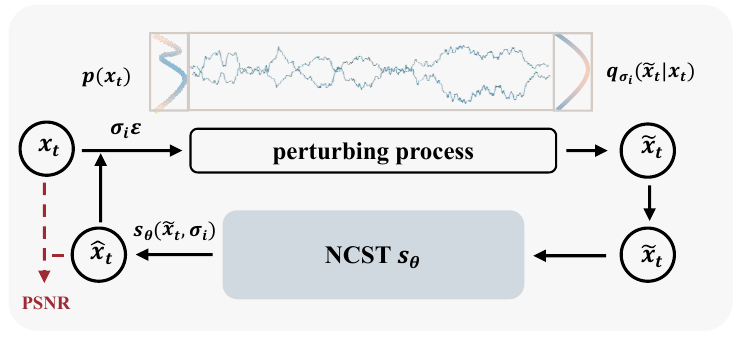}
  \caption{ An overview of the proposed autoregressive denoising score matching process.}
  \label{fig:fig5}
\end{figure}

\subsection{Autoregressive Denoising Mechanism for Anomaly Detection}

To amplify responses to local-mode anomalies while preserving low scores for normal data, we propose an autoregressive DSM (ADSM) mechanism across multiple noise levels. As illustrated in Fig.~\ref{fig:fig5}, for each perturbed sample $\tilde{x}_t$ obtained by adding noise $\sigma_i$ to the original sequence, the denoised output is computed as $\hat{x}_t = \tilde{x}_t + \sigma_i^2 s_\theta(\tilde{x}_t, \sigma_i)$. Crucially, we improve the mechanism by autoregressively applying intensifying noise to the denoised data instead of the original data, and estimate the score at each step. This autoregressive refinement accumulates the abnormal context over iterations, yielding a more effective score function.

Furthermore, to bridge the appearance gap, we incorporate visual fidelity via PSNR between the original and denoised frames. Specifically, for each denoising step at noise level $\sigma_i$, we define the per-noise-level anomaly score as:
\begin{equation}
    \text{score}_i(t) = \frac{ \| s_\theta(\tilde{x}_t, \sigma_i) \| }{PSNR(\hat{x}_t, x_t) }
\end{equation}
where the denominator acts as a visual consistency regularizer. When likelihood and appearance agree (e.g., normal data), the score remains low. In contrast, when they disagree (e.g., anomalous motion with plausible appearance), the denominator suppresses false positives while the numerator compensates for missed detections.

\subsection{Self-Distilled Denoising Mechanism for Anomaly Anticipation}
To enable ground-truth-free supervision for video anomaly anticipation (VAA), we propose a novel self-distilled DSM (SDSM) mechanism. Unlike conventional knowledge distillation, SDSM leverages the consistency of score estimates under the learned normal distribution. Specifically, since both the main and auxiliary branches are trained exclusively on normal data, they produce highly consistent denoised outputs for normal test samples. In contrast, anomalous inputs tend to induce noticeable divergence between their outputs. Consequently, the discrepancy between the two predictions provides a reliable proxy for anomaly probability, circumventing the absence of future ground truth in VAA.

As illustrated in Fig.~\ref{fig:fig3}, SDSM is built upon the noise-conditioned score transformer (NCST). It comprises a shared backbone with two branches, including a main branch and an auxiliary branch. The auxiliary branch diverges after the first transformer block of the main branch and adds only one extra transformer layer.

Specifically, let $x_{\text{in}}^t = \{f_{t-n}, \dots, f_{t-1}\}$ denote the input video clip at time $t$, and $x_{\text{tar}}^t = \{f_{t+1}, \dots, f_{t+n}\}$ denote the target video clip at time $t$. Let $\{\sigma_i\}_{i=1}^L$ be a geometric noise schedule satisfying $\sigma_L / \sigma_{L-1} = \cdots = \sigma_2 / \sigma_1 > 1$. 
At noise level $\sigma_i$, the input and target video sequences are perturbed with Gaussian noise $\boldsymbol{\epsilon} \sim \mathcal{N}(0, \sigma_i^2 I)$ to obtain $\tilde{x}_{\text{in}}^t$ and $\tilde{x}_{\text{tar}}^t$, where $q_{\sigma_i}(\tilde{x} | x) = \mathcal{N}(\tilde{x} | x, \sigma_i^2 I)$. 
We partition the $n$ perturbed frames from the input sequence $\tilde{x}_{\text{in}}^t$ and the $n$ perturbed frames from the target sequence $\tilde{x}_{\text{tar}}^t$ using the uniform frame patch embedding method, embedding each individual frame into tokens. 
Consequently, we obtain tokens $\{\tilde{P}_{in}^j\}_{j=1}^N \in \mathbb{R}^{d \times d \times c}$, $\{P_{in}^j\}_{j=1}^N \in \mathbb{R}^{d \times d \times c}$, $\{\tilde{\mathbf{p}}_{tar}^j\}_{j=1}^N \in \mathbb{R}^{d \times d \times c}$, and $\{\mathbf{p}_{tar}^j\}_{j=1}^N \in \mathbb{R}^{d \times d \times c}$ from $\tilde{x}_{\text{in}}^t$, $x_{\text{in}}^t$, $\tilde{x}_{\text{tar}}^t$, and $x_{\text{tar}}^t$, respectively. 
Here, $N$ denotes the number of non-overlapping patches of size $d \times d \times c$ per sequence, where $c$ is the number of input channels and $d$ is a hyperparameter that determines the relationship between $N$ and $n$. 
As described in Section~\ref{sec:dsm}, the noise-conditional score transformer model $s_\theta(\cdot, \sigma)$ is trained via denoising score matching to estimate the score function $\nabla_{\tilde{x}} \log q_{\sigma}(\tilde{x})$ for noisy patches. 
This is achieved by minimizing the following weighted score-matching loss over $L$ noise levels and $N$ patches per sequence:
\begin{equation}
\begin{aligned}
\mathcal{L}_{\text{main}} ={}& \frac{1}{L} \sum_{i=1}^{L} \frac{\sigma_i^2}{2}
\sum_{j=1}^{N} \omega_j \,
\mathbb{E}_{\,q_{\sigma_i}(\tilde{P}_{in}^j \mid x)\,p(x)} \\
&\hspace{6em} \left\| s_\theta(\tilde{P}_{in}^j, \sigma_i) + \frac{\tilde{\mathbf{p}}_{tar}^j - \mathbf{p}_{tar}^j}{\sigma_i^2} \right\|^2
\end{aligned}
\label{eq:loss_function1}
\end{equation}

To enhance motion sensitivity, we reuse the key-frame motion weighting scheme from Section~\ref{subsec:motion_weighting}, which remains applicable here because anomaly cues in both detection and anticipation tasks are strongly linked to motion variations. For each input sequence $x_{\text{in}}$, the motion saliency weights $\omega_j$ are computed according to Eq.~\eqref{eq:gradient_diff}, Eq.~\eqref{eq:channel_pooling}, and Eq.~\eqref{eq:omega_weight}. Reweighting each patch-level term with $\omega_j$ emphasizes dynamic regions while suppressing static background, improving motion sensitivity and anomaly discriminability.

Training proceeds in two stages. In the first stage, the main decoder is trained using the motion-aware objective defined in Eq.~\eqref{eq:loss_function1}. In the second stage, the shared backbone (i.e., the encoder and main decoder) is frozen, and the auxiliary decoder is trained via self-distillation to reconstruct the denoised target patches $\hat{\mathbf{p}}_{tar}^j = \tilde{\mathbf{p}}_{tar}^j + \sigma_i^2 s_\theta(\tilde{\mathbf{p}}_{tar}^j, \sigma_i)$ generated by the main decoder. Let $\{\bar{\mathbf{p}}_{tar}^j\}_{j=1}^N$ denote the output of the auxiliary decoder. The self-distillation loss is formulated as:
\begin{equation}
    \mathcal{L}_{\text{SD}} = \frac{1}{L} \sum_{i=1}^{L} \left\{ \sum_{j=1}^{N} \omega_j \cdot \left\| \bar{\mathbf{p}}_{tar}^j - \hat{\mathbf{p}}_{tar}^j \right\|_2^2 \right\}
\end{equation}
where $\bar{\mathbf{p}}_{tar}^j$ and $\hat{\mathbf{p}}_{tar}^j$ represent the patches denoised by the auxiliary and main branches, respectively. The motion weight $\omega_j$ is reused from Eq.~\eqref{eq:omega_weight}.

As an integral component of the unified Uni-DSM framework illustrated in Fig.~\ref{fig:fig3}, we employ the shared noise-conditioned score transformer (NCST) backbone to drive the VAA task. Leveraging its attention-based mechanisms, this architecture effectively constructs a stable vector field through consecutive supervision of the multiscale patch-wise denoising process, thereby making anomalies clearly distinguishable in video action anticipation (VAA) tasks.

\subsection{Score-Based Anomaly Inference}
In both video anomaly detection (VAD) and video anomaly anticipation (VAA), anomaly inference is fundamentally grounded in the learned data density. Since our model estimates the score function $s_\theta(x, \sigma) = \nabla_{x} \log p(x)$, anomaly evidence can be directly derived from the magnitude of the log-density gradient:
\begin{equation}
    \|s_\theta(x)\| = \|\nabla_{x} \log p(x)\| = \left\| \frac{\nabla_{x} p(x)}{p(x)} \right\|
\end{equation}

Intuitively, samples that lie in high-density regions corresponding to normal events tend to produce small score magnitudes, whereas samples in low-density regions corresponding to anomalous events yield stronger responses.

\begin{algorithm}[t]
\caption{Autoregressive Denoising Score Matching for VAD}
\label{alg:infer}
\begin{algorithmic}[1]
\Require Number of noise levels $L$; \\
$\epsilon \gets \text{sample } \mathcal{N}(0, I)$; \\
$\sigma \sim \text{sample log-uniform}(\sigma_1, \sigma_L)$; \\
Input video sequence $x_t = \{f_{t-n}, \dots, f_{t-1}\}$; \\
Score network $s_\theta$
\Ensure Anomaly score $S(t)$
\State $x_t = \{f_{t-n}, \dots, f_{t-1}\}$
\State $\hat{x}_t \gets x_t$
\For{$i = 1$ to $L$}
    \State $\tilde{x}_t \gets \hat{x}_t + \sigma_i \epsilon$
    \State $\hat{x}_t \gets \tilde{x}_t + \sigma_i^2 s_\theta(\tilde{x}_t, \sigma_i)$
    \State $\text{score}_i(t) \gets \| s_\theta(\tilde{x}_t, \sigma_i) \| / \text{PSNR}(\hat{x}_t, x_t)$
\EndFor
\State $S(t) \gets \text{normalize}\big(\{\text{score}_i(t)\}_{i=1}^L\big)$
\State \Return $S(t)$
\end{algorithmic}
\end{algorithm}

In VAD, the entire video sequence is observable, enabling direct score evaluation on each frame. 
Given a test video, we divide it into $\eta$ segments of $T$ frames each. For segment $t$, we compute $L$ scores $\{\text{score}_i(t)\}_{i=1}^L$ across noise levels and take the maximum per level to enhance separability~\cite{yang2023video, cao2024scene}:
\begin{equation}
\text{score}_i(t) = \max_{1 \leq \tau \leq T} \text{score}_i(\tau)
\end{equation}

Each  $\text{score}_i(t)$ is then normalized to $[0,1]$ to obtain the calibrated anomaly probability~\cite{mathieu2015deep}:
\begin{equation}
\begin{split}
    S_i(t) &= \text{normalize}\big(\text{score}_i(t)\big) \\
           &= \frac{ \text{score}_i(t) - \min\limits_{1\leq \tau \leq \eta T} \text{score}_i(\tau) }
                  { \max\limits_{1\leq \tau \leq \eta T} \text{score}_i(\tau) - \min\limits_{1\leq \tau \leq \eta T} \text{score}_i(\tau) }
\end{split}
\end{equation}

Finally,  we weigh and average these anomaly scores $S_i(t)$ to obtain an ultimate anomaly indicator, which is strong enough for VAD. This yields a robust, multi-scale anomaly indicator suitable for VAD, as summarized in Algorithm~\ref{alg:infer}.

Unlike VAD, VAA operates under partial observability where future frames are unavailable at inference time. Consequently, direct comparison with future ground truth cannot be directly applied. 
Nevertheless, the score function remains well-defined over the learned distribution. Therefore, we retain the score norm normalized by visual fidelity as the primary anomaly metric.

However, an additional challenge arises in anticipation because the score function inherently approaches zero near local density modes. In detection, such cases can often be compensated by reconstruction discrepancy, whereas in anticipation, no future ground truth is available to provide this corrective signal. As a result, anomalies residing in local modes may remain undetected. To address this limitation while preserving the probabilistic foundation of inference, we leverage the SDSM mechanism. For each perturbed input $\tilde{x}^{t}_{\text{in}}$ and target $\tilde{x}^{t}_{\text{tar}}$ at noise level $\sigma_i$, the main decoder generates the denoised target $\hat{x}^{t}_{\text{tar}} = \tilde{x}^{t}_{\text{tar}} + \sigma_i^2 s_\theta(\tilde{x}^{t}_{\text{in}}, \sigma_i)$, while the auxiliary decoder reconstructs $\bar{x}_t^{\text{tar}}$ from $\hat{x}^{t}_{\text{tar}}$. The discrepancy between their outputs serves as a surrogate for anomaly probability. Crucially, in each denoising step, we use the PSNR between $\hat{x}^{t}_{\text{tar}}$ and $\bar{x}_t^{\text{tar}}$ as the denominator of the score norm, yielding the anomaly score:
\begin{equation}
    \text{score}_i(t) = \frac{ \| s_\theta(\tilde{x}^{t}_{\text{in}}, \sigma_i) \| }{ PSNR(\bar{x}_t^{\text{tar}}, \hat{x}_{tar}^t) } 
\end{equation}

When the likelihood and appearance cues in the numerator and denominator are consistent, the model yields highly confident and reliable predictions for both normal and abnormal events. In contrast, when they diverge, the regions where the numerator misjudges are suppressed by the denominator, while the parts overlooked by the denominator are compensated by the numerator, resulting in a more robust and balanced score with the appearance gap filled.

For the VAA task, let $\alpha$ denote the anticipation horizon. We treat $\text{score}_i(t+k)$ as the anomaly score for the $k$-th future frame ($k \in [1, \alpha]$). The predicted anomaly score for the window $[t+1, t+\alpha]$ is then defined as the maximum over the $\alpha$ frames:
\begin{equation}
\text{score}_i(t+1:t+\alpha) = \max_{1 \leq k \leq \alpha} \text{score}_i(t+k)
\end{equation}

Each $\text{score}_i(t+1:t+\alpha)$ is normalized to $[0,1]$ over the temporal window $\tau \in [t+1, t+\alpha]$ to obtain the calibrated anomaly probability:
Each $\text{score}_i(t)$ is then normalized to $[0,1]$ to obtain the calibrated anomaly probability~\cite{mathieu2015deep}:
\begin{equation}
\begin{split}
    S_i(t+1:t+\alpha) 
    &= \text{normalize}\big(\text{score}_i(t+1:t+\alpha)\big) \\
    &\hspace*{-5em} = \frac{
               \text{score}_i(t+1:t+\alpha) - \min\limits_{t+1 \leq \tau \leq t+\alpha} \text{score}_i(\tau)
             }{
               \max\limits_{t+1 \leq \tau \leq t+\alpha} \text{score}_i(\tau) - \min\limits_{t+1 \leq \tau \leq t+\alpha} \text{score}_i(\tau)
             }.
\end{split}
\end{equation}

Finally, the overall anomaly probability for the anticipation window is obtained by computing a weighted average of the normalized scores across all noise levels. The entire process is detailed in Algorithm \ref{alg:vaa}.

\begin{algorithm}[t]
\caption{Self-Distillation Denoising Score Matching for VAA}
\label{alg:vaa}
\begin{algorithmic}[1]
\Require 
Number of noise levels $L$; \\
$\epsilon \gets \text{sample } \mathcal{N}(0, I)$; \\
$\sigma \sim \text{sample log-uniform}(\sigma_1, \sigma_L)$; \\
Input video sequence $x_{\text{in}}^t = \{f_{t-n}, \dots, f_{t-1}\}$; \\
Target video sequence $x_{\text{tar}}^t = \{f_{t+1}, \dots, f_{t+n}\}$; \\
Main score network $s_\theta$; \\
Sub score network $s_\theta'$
\Ensure 
Anomaly score $S_i(t+1: t+\alpha)$
\State $x_{\text{in}}^t = \{f_{t-n}, \dots, f_{t-1}\}$
\State $x_{\text{tar}}^t = \{f_{t+1}, \dots, f_{t+n}\}$
\For{$i = 1$ to $L$}
    \State $\tilde{x}_{\text{in}}^t \gets x_{\text{in}}^t + \sigma_i \epsilon$
    \State $\hat{x}_{tar}^t \gets \tilde{x}_{\text{tar}}^t + \sigma_i^2 s_\theta(\tilde{x}_{\text{in}}^t, \sigma_i)$
    \State $\bar{x}_{\text{tar}}^t \gets s_\theta'(\tilde{x}_{\text{in}}^t, \sigma_i)$
    \State $\text{score}_i(t) \gets \| s_\theta(\tilde{x}_t^{\text{in}}, \sigma_i) \| / \text{PSNR}(\bar{x}_t^{\text{tar}}, \hat{x}_{tar}^t)$
\EndFor
\State $\text{score}_i(t+1: t+\alpha) \gets \max(\text{score}_i(t+k))$
\State $S_i(t+1: t+\alpha) \gets \text{normalize}(\text{score}_i(t+1: t+\alpha))$
\State \Return $S_i(t+1: t+\alpha)$
\end{algorithmic}
\end{algorithm}

\section{Experiments}
\subsection{Experimental Setup}
\textbf{Datasets.} To comprehensively evaluate our method, we utilize the CUHK Avenue dataset~\cite{lu2013abnormal}, the ShanghaiTech dataset~\cite{luo2017revisit} and the NWPU Campus dataset~\cite{cao2023new}. 
The CUHK Avenue dataset (hereafter referred to as the Avenue dataset)~\cite{lu2013abnormal} is a widely used benchmark containing 16 training videos and 21 test videos, totaling 35,240 frames, with each video lasting approximately two minutes. Typical anomalies include running on sidewalks, throwing bags, and children jumping rope. 
The ShanghaiTech dataset~\cite{luo2017revisit} consists of 330 training videos and 107 test videos captured across 13 distinct scenes, with anomalies such as fighting, robbery, and brawling. 
The NWPU Campus dataset (hereafter referred to as the NWPU dataset)~\cite{cao2023new} is currently the largest semi-supervised video anomaly detection benchmark, comprising 305 training videos and 242 test videos with 1,466,073 frames in total. It covers 43 different scenes with anomalies including jaywalking and illegal U-turns. Notably, it is the first dataset that explicitly considers scene-dependent anomalies and supports video anomaly anticipation (VAA).

\textbf{Evaluation Metric.} We adopt the Area Under the ROC Curve (AUC) metric for both VAD and VAA. In VAD, we evaluate performance using both macro-average and micro-average AUC~\cite{acsintoae2022ubnormal}, \cite{barbalau2023ssmtl++}, \cite{georgescu2021background}, \cite{reiss2022attribute}. Macro-AUC computes the AUC for each test video individually and then averages across all videos, while Micro-AUC concatenates all frames from all test videos before computing a single AUC. For VAA, we concatenate all frames in the dataset and compute frame-level AUC.

\textbf{Implementation details.} Input frames $x_t$ are cropped to $160 \times 160$ pixel regions centered on objects detected by the pre-trained ByteTrack~\cite{zhang2022bytetrack}, which is implemented in MMTracking~\cite{mmtrack}. This cropping strategy enables the model to focus on salient targets while implicitly capturing contextual scene information. If a frame contains multiple targets, each is processed independently. For VAD, our model $s_\theta$ takes an 8-frame input sequence, partitioned into $N = 8 \times 100$ non-overlapping patches $\{\mathbf{P}_j\}_{j=1}^N \in \mathbb{R}^{d \times d \times c}$ with patch size $d = 16$. The model is trained for 100 epochs with a batch size of 20 using the Adamax optimizer~\cite{kingma2014adam}. The initial learning rate is set to $10^{-4}$ and gradually decays following the scheme of cosine annealing. During training, the noise scale $\sigma_i$ for each batch element is sampled from a log-uniform distribution over $[\sigma_1, \sigma_L] = [0.001, 1.0]$. During inference, we use $L = 20$ uniformly spaced noise levels within this interval. Both scene category embeddings and diffusion time embeddings have dimensions of 768 and are concatenated for model learning and inference. For VAA, the input sequence length is reduced to 5 frames ($N = 5 \times 100$ patches, $d = 16$), and the model is trained for 50 epochs, while all other settings remain unchanged. All experiments are conducted on four NVIDIA GeForce RTX 4090 GPUs using the PyTorch~\cite{paszke2019pytorch} framework.

\begin{table}[t]
\small 
\caption{Comparison of different methods on the Avenue, ShanghaiTech, and NWPU datasets in AUCs (\(\%\)) metric. The best result on each dataset is shown in bold. We also list the inputs of the methods, in which method without marking denotes single modal input.}
\setlength\tabcolsep{2pt} 
\centering
\begin{tabularx}{\columnwidth}{@{}l*{6}{>{\centering\arraybackslash}X}@{}}
\hline
\multirow{2}{*}{Method} & \multicolumn{2}{c}{Avenue} & \multicolumn{2}{c}{ShanghaiTech} & \multicolumn{2}{c}{NWPU} \\ \cline{2-7} 
                        & Micro & Macro & Micro & Macro & Micro & Macro \\ \hline
\(^\Diamond\)FFP~\cite{liu2018future}                     & 84.9 & -           & 72.8 & -              & -    & - \\
\(^\Diamond\)AMMC-Net~\cite{cai2021appearance}            & 86.6 & -           & 73.7 & -              & 64.5 & - \\
\(^\Diamond\)HF\(^2\)-VAD~\cite{liu2021hybrid}            & 91.1 & \underline{93.5} & 76.2 & 83.1 & 63.7 & - \\
\(^\Diamond\)VABD~\cite{li2021variational}                & 86.6 & -           & 78.2 & -              & -    & - \\ \hline
\(^+\)MTP~\cite{rodrigues2020multi}                       & 82.9 & -           & 76.0 & -              & -    & - \\
\(^+\)STG-NF~\cite{hirschorn2023normalizing}              & -    & -           & \textbf{85.9} & -     & -    & - \\
\(^+\)MoCoDAD~\cite{flaborea2023multimodal}               & 89.0 & -           & 77.9 & -              & -    & - \\ \hline
MemAE~\cite{gong2019memorizing}                           & 84.9 & -           & 72.8 & -              & 61.9 & - \\
CAE~\cite{ionescu2019object}                              & 87.4 & 90.4        & 78.7 & 84.9           & -    & - \\
MNAD~\cite{park2020learning}                              & 88.5 & -           & 70.5 & -              & 62.5 & - \\
OG-Net~\cite{zaheer2020old}                               & -    & -           & -    & -              & 62.5 & - \\
MPN~\cite{lv2021learning}                                 & 89.5 & -           & 73.8 & -              & 64.4 & - \\
SSMTL~\cite{georgescu2021anomaly}                         & \underline{91.5} & 91.9 & 82.4 & 89.3 & - & - \\
LLSH~\cite{lu2022learnable}                               & 87.4 & 88.6        & 77.6 & 85.9           & 62.2 & - \\
FBAE~\cite{cao2023new}                                    & 86.8 & -           & 79.2 & -              & 68.2 & - \\
FPDM~\cite{yan2023feature}                                & 90.1 & -           & 78.6 & -              & -    & - \\
Two-stream~\cite{cao2024context}                          & 90.8 & 93.0        & 83.7 & \underline{90.8} & - & - \\
SDMAE~\cite{ristea2024self}                               & 91.3 & 90.9        & 79.1 & 84.7           & -    & - \\
SSAE~\cite{cao2024scene}                                  & 90.2 & 91.5        & 80.5 & 88.9           & \underline{75.6} & \underline{76.9} \\ \hline
Uni-DSM (ours)                                            & \textbf{91.6} & \textbf{94.2} & \underline{84.5} & \textbf{93.2} & \textbf{76.9} & \textbf{78.1} \\ \hline
\end{tabularx}
\label{tab:performance}
\end{table}
\begin{table}[t]
\caption{Anticipation Results of different methods on the NWPU Dataset}
\centering
\begin{tabular*}{\columnwidth}{@{\extracolsep{\fill}} l c c c c c c @{}}
\hline
$\alpha$ & 0.5s & 1s & 1.5s & 2s & 2.5s & 3s \\
\hline
Chance & 50.0 & 50.0 & 50.0 & 50.0 & 50.0 & 50.0 \\
Human & -- & -- & -- & -- & -- & \textcolor{gray}{90.4} \\
FBAE~\cite{cao2023new} & 65.8 & 65.3 & 64.9 & 64.6 & 64.2 & 64.0 \\
SSAE~\cite{cao2024scene} & \underline{69.5} & \underline{69.3} & \underline{68.7} & \underline{68.2} & \underline{67.1} & \underline{66.9} \\
Uni-DSM (ours) & \textbf{71.3} & \textbf{71.1} & \textbf{70.5} & \textbf{70.2} & \textbf{69.5} & \textbf{69.1} \\
\hline
\end{tabular*}
\label{tab:SDSM_results}
\end{table}
\subsection{VAD Performance Benchmarking}
Table~\ref{tab:performance} presents the VAD performance comparison between our method and existing approaches on the Avenue~\cite{lu2013abnormal}, ShanghaiTech~\cite{luo2017revisit}, and NWPU Campus~\cite{cao2023new} datasets. The results are reported using both the micro and macro scores. We also list the input modalities of different methods, where unmarked entries denote single-modality inputs, ``$\diamond$'' and ``$+$'' indicate multimodal inputs incorporating optical flow and skeleton data, respectively. We prefer the macro score, especially on multi-scene datasets like ShanghaiTech~\cite{luo2017revisit} and NWPU~\cite{cao2023new}, as ambiguous scene boundaries prevent micro score from comprehensively and fairly judging the performance. On all three datasets, each containing more than 10 anomaly categories, our method outperforms all competitors in terms of macro score. Notably, we achieve a significant improvement of up to 2.4\% on the ShanghaiTech dataset~\cite{luo2017revisit}, demonstrating effectiveness of our method. Moreover, on the NWPU dataset~\cite{cao2023new}, our method attains the best performance on both metrics, validating the scalability and stability of our method. Compared with multimodal-input methods, our approach, using only raw video clips, achieves performance that is not only comparable but often superior. Against methods employing the same single-modality input, our method consistently achieves state-of-the-art results across all three datasets, obtaining the highest scores in both micro and macro evaluations. This outstanding performance confirms that our Uni-DSM can effectively detect anomalies in complex and challenging video anomaly detection tasks. It is worth noting that since the score is computed based on random noise, it inherently possesses a certain degree of uncertainty. 
However, the qualitative analysis in Section~\ref{sec:qualitative} verifies the validity of the denoising score matching mechanism. By fusing conditional embeddings with visual features, our method  maximally controls this uncertainty. 
Furthermore, extensive testings reveal minimal performance fluctuations ($<$1\% across multiple runs), confirming the stability and reliability of our reported results.

\subsection{VAA Performance Benchmarking}
To evaluate the effectiveness of our self-distilled DSM (SDSM) mechanism on the VAA task, we follow the standard video anomaly anticipation protocol on the NWPU Campus dataset~\cite{cao2023new}, which requires predicting anomalies that occur within a specific future time window. Comparative experiments are conducted across different anticipation horizons. Tables~\ref{tab:SDSM_results} present the performance of the SDSM on the NWPU dataset~\cite{cao2023new}, with a focus on comparing different methods in terms of anticipation accuracy. We also report the results of stochastic anticipations (“Chance”) and human beings (“Human”). Four volunteers not involved in the construction of the dataset participate in the evaluation of anomaly anticipation. Since humans cannot perceive time precisely, the volunteers only anticipate whether an anomalous event will occur in 3 seconds or not. The result of “Human” is the average performance of all the volunteers. For algorithms, there is still much room for improvement, indicating that the VAA task is very challenging. Notebly, SDSM consistently outperforms both the random baseline and existing learning-based approaches across all anticipation horizons from 0.5s to 3s. 
Compared with the forward-backward frame prediction framework FBAE~\cite{cao2023new}, SDSM achieves a clear improvement of around 5\%, while maintaining stable performance as the anticipation horizon increases. 
Furthermore, compared with the stronger bidirectional modeling approach SSAE~\cite{cao2024scene}, SDSM still yields consistent gains across all horizons, with an average improvement of 2.1\%. 
Although performance naturally declines as the anticipation horizon extends, SDSM consistently outperforms all baselines and remains computationally efficient, with detailed efficiency analysis provided in Section~\ref{sec:computational}.

\subsection{Study on Latent Space}
Latent-space generative models are widely used to remove redundancy from raw inputs and improve efficiency in synthesizing high-quality images and videos~\cite{ma2024latte}, \cite{rombach2022high}, \cite{bao2023all}, \cite{bao2023one}. However, VAD differs from typical generative tasks in that it is highly sensitive to the underlying data distribution, implying limited redundancy in the input. Consequently, projecting data into a compact latent space may discard subtle but critical cues for anomaly detection. 
These observations highlight the importance of directly modeling raw video signals. In practice, improvements in VAD performance often stem from stronger feature representations rather than more sophisticated likelihood modeling schemes. This motivates us to reconsider whether performing denoising score matching in the latent space of pre-trained generative models is necessary for VAD.

We adopt the variational autoencoder used in Stable Diffusion~\cite{rombach2022high}, pre-trained on the COCO dataset~\cite{lin2014microsoft}, to encode clean input sequences into latent features. Our ADSM, instantiated with the NCST shown in Fig.~\ref{fig:fig3}, is then applied within this latent representation space. Quantitative results in Table~\ref{tab:latent}, evaluated on the ShanghaiTech~\cite{luo2017revisit} and NWPU~\cite{cao2023new} benchmarks using both micro and macro metrics, show that operating in latent space yields inferior performance compared to modeling original data directly. This supports our hypothesis that the inconsistent distribution between the features extracted by the pre-trained model and the VAD dataset exacerbates the unstable performance of the latent version. Although representation alignment techniques~\cite{yu2024representation} could alleviate this issue, directly implementing our ADSM on the original data space is more intuitive with the advancement of computing resources and model capabilities.

\begin{table}[t]
\caption{AUCs (\(\%\)) of our model in latent space and original data space on ShanghaiTech and NWPU datasets.}
\centering
\begin{tabular*}{0.95\columnwidth}{@{\extracolsep{\fill}} l c c c c @{}}
\hline
\multirow{2}{*}{Method} & \multicolumn{2}{c}{ShanghaiTech} & \multicolumn{2}{c}{NWPU} \\ \cline{2-5}
                        & Micro & Macro & Micro & Macro \\ \hline
L-ADSM                  & 77.6  & 87.5  & 68.1  & 73.2  \\
Uni-DSM                    & \textbf{84.5} & \textbf{93.2} & \textbf{76.9} & \textbf{78.1} \\ \hline
\end{tabular*}
\label{tab:latent}
\end{table}

\begin{table}[t]
\caption{AUCs (\(\%\)) of different likelihood-based methods in the original data space on ShanghaiTech and NWPU datasets.}
\centering
\begin{tabular*}{0.95\columnwidth}{@{\extracolsep{\fill}} l c c c c @{}}
\hline
\multirow{2}{*}{Method} & \multicolumn{2}{c}{ShanghaiTech} & \multicolumn{2}{c}{NWPU} \\ \cline{2-5}
                        & Micro          & Macro          & Micro       & Macro      \\ \hline
MSMA~\cite{mahmood2020multiscale} & 81.3 & \underline{90.4} & 72.3 & \underline{75.1} \\
MULDE~\cite{micorek2024mulde}     & \underline{82.6} & \underline{90.4} & \underline{73.1} & 74.5 \\
Uni-DSM                              & \textbf{84.5} & \textbf{93.2} & \textbf{76.9} & \textbf{78.1} \\ \hline
\end{tabular*}
\label{tab:mulde}
\end{table}

\begin{table}[t]
\caption{AUCs(\(\%\)) of our model variants on the ShanghaiTech dataset for the VAD task.}
\centering
\renewcommand{\arraystretch}{1.1}
\begin{tabularx}{\linewidth}{
c|
>{\centering\arraybackslash}p{0.08\linewidth}
>{\centering\arraybackslash}p{0.08\linewidth}
>{\centering\arraybackslash}p{0.08\linewidth}
>{\centering\arraybackslash}p{0.10\linewidth}
>{\centering\arraybackslash}p{0.14\linewidth}|
>{\centering\arraybackslash}X
>{\centering\arraybackslash}X}
\hline
\multirow{2}{*}{} & \multirow{2}{*}{DSM} & \multirow{2}{*}{ADSM} & 
\multirow{2}{*}{Scene} & \multirow{2}{*}{Motion} & 
\multirow{2}{*}{Appearance} & 
\multicolumn{2}{c@{}}{ShanghaiTech} \\
\cline{7-8}
 & & & & & & Micro & Macro \\
\hline
1 & \checkmark &  &  &  &  & 73.9 & 78.5 \\
2 & \checkmark & \checkmark &  &  &  & 76.4 & 84.1 \\
3 & \checkmark & \checkmark & \checkmark &  &  & 76.9 & 86.3 \\
4 & \checkmark & \checkmark &  & \checkmark &  & 78.6 & 87.7 \\
5 & \checkmark & \checkmark &  &  & \checkmark & \underline{82.1} & \underline{89.9} \\
6 & \checkmark & \checkmark & \checkmark & \checkmark & \checkmark & \textbf{84.5} & \textbf{93.2} \\
\hline
\end{tabularx}
\label{tab:ablationVAD}
\end{table}

\begin{table}[t]
\caption{AUCs(\%) of our model variants on the NWPU Dataset for the VAA task.}
\centering
\setlength{\tabcolsep}{2pt}

\begin{tabularx}{0.95\linewidth}{l|*{4}{>{\centering\arraybackslash}X}|c}
\hline
 & DSM & SDSM & Scene & Motion & NWPU datasets \\
\hline
1 & \checkmark &  &  &  & 67.5 \\
2 & \checkmark & \checkmark &  &  & 68.1 \\
3 & \checkmark & \checkmark & \checkmark &  & 68.6 \\
4 & \checkmark & \checkmark &  & \checkmark & \underline{69.3} \\
5 & \checkmark & \checkmark & \checkmark & \checkmark & \textbf{70.5} \\
\hline
\end{tabularx}
\label{tab:ablationVAA}
\end{table}

To further investigate this issue, we examine two likelihood-based baselines closely related to our framework, including a latent-space method~\cite{micorek2024mulde} and an approach originally proposed for image anomaly detection~\cite{mahmood2020multiscale}. For fair comparison, both methods are re-implemented under consistent experimental settings and evaluated on the ShanghaiTech and NWPU benchmarks. Results in Table~\ref{tab:mulde} show that our method consistently outperforms these baselines, demonstrating that applying denoising score matching in the original data space is more effective. The original data space enables us to take both the efficiency of likelihood estimation and the characteristic of video anomalies into account, further supporting our motivation.

\subsection{Ablation Studies}
\label{sec:ablation}
To further verify the effectiveness of each specific design of Uni-DSM, we conduct ablation experiments to compare Uni-DSM with its several model variants. Table~\ref{tab:ablationVAD} reports the detection performance of model variants on the ShanghaiTech dataset~\cite{luo2017revisit}. Components are progressively added to the noise-conditional score transformer, including denoising score matching (DSM), autoregressive denoising score matching (ADSM), scene-conditioned embedding (Scene), key-frame motion weighting (Motion), and visual difference aggregation (Appearance). As shown in Table~\ref{tab:ablationVAD}, the autoregressive mechanism significantly improves DSM for visual difference aggregation in VAD, yielding an average gain of 4.1\% in both micro and macro scores. Although ShanghaiTech does not explicitly emphasize scene-dependent anomalies, the macro score still increases by 2.2\%, suggesting that ADSM captures scene-aware information more effectively. Introducing motion consistency further improves performance by 2.9\%, indicating that ADSM effectively associates spatiotemporal information. Aggregating the discrepancy between denoised and original data proves to be the most effective strategy for constructing a discriminative anomaly metric, improving micro and macro scores by 5.7\% and 5.8\%, respectively. When all components are combined, the micro and macro scores further increase by 2.4\% and 3.3\%, respectively, demonstrating strong complementarity among the modules of ADSM.

Table~\ref{tab:ablationVAA} reports the anticipation performance of model variants on the NWPU dataset~\cite{cao2023new}. Components are progressively added to the noise-conditional score transformer, including denoising score matching (DSM), self-distilled denoising score matching (SDSM), scene-conditioned embedding (Scene), and key-frame motion weighting (Motion) to validate their contributions. Each module improves performance. The DSM baseline achieves 67.5\% accuracy. Adding SDSM increases accuracy to 68.1\%, indicating that self-distillation optimizes denoising and enhances anticipation. Scene-conditioned embedding raises performance to 68.6\%, highlighting the auxiliary role of scene context. Key-frame motion weighting further increases accuracy to 69.3\%, emphasizing critical motion cues for discriminative temporal features. When all modules are integrated, the model achieves 70.5\%, consistently outperforming both individual modules and their partial combinations. This demonstrates strong complementarity among the modules, resulting in a more robust and accurate anomaly anticipation mechanism.

In summary, the ablation studies systematically validate the effectiveness of each proposed component. Their incremental integration consistently yields substantial performance gains, and the full combination achieves the best results, validating the effectiveness of our model architecture.

\begin{table}[t]
\caption{Comparison of computational resource consumption during VAD inference on the ShanghaiTech dataset.}
\label{tab:shanghaitech_resource}
\centering
\setlength{\tabcolsep}{4pt}
\renewcommand{\arraystretch}{1.1}
\begin{tabularx}{\columnwidth}{l c c c c}
\toprule
Model 
& \makecell{GPU\\Memory (G)} 
& \makecell{Running\\Speed (s)} 
& \makecell{Frame Rate\\(FPS)} 
& \makecell{Performance\\(\%)} \\
\midrule
$^\diamond$HF$^2$-VAD~\cite{liu2021hybrid} & -- & 0.0667 & 15.0 & 76.2 \\
$^\diamond$VABD~\cite{li2021variational} & -- & \textbf{0.0147} & \textbf{68.0} & 78.2 \\
LLSH~\cite{lu2022learnable}              & -- & 0.0392 & 25.5 & 77.6 \\
FPDM~\cite{yan2023feature}               & -- & 0.1282 & 7.8  & 78.6 \\
SSAE~\cite{cao2024scene}                 & \underline{24} & 0.1096 & 10.1 & \underline{80.5} \\
Uni-DSM (ours)                     & \textbf{11} & \underline{0.0194} & \underline{51.5} & \textbf{84.5} \\
\bottomrule
\end{tabularx}
\end{table}

\begin{table}[t]
\caption{Comparison of Computational Resource Consumption During VAA Inference on the NWPU Dataset}
\label{tab:nwpu_resource}
\centering
\setlength{\tabcolsep}{4pt}
\renewcommand{\arraystretch}{1.1}
\begin{tabularx}{\columnwidth}{l c c c c}
\toprule
Model 
& \makecell{GPU\\Memory (G)} 
& \makecell{Running\\Speed (s)} 
& \makecell{Frame Rate\\(FPS)} 
& \makecell{Performance\\(\%)} \\
\midrule
EPAP-Net~\cite{leng2022anomaly} & -- & \underline{0.0219} & \underline{45.7} & -- \\
FBAE~\cite{cao2023new}          & \underline{21} & 0.0274 & 36.5 & 64.9 \\
SSAE~\cite{cao2024scene}        & 24 & 0.1096 & 10.1 & \underline{68.7} \\
SDSM (ours)            & \textbf{16} & \textbf{0.0163} & \textbf{61.3} & \textbf{70.5} \\
\bottomrule
\end{tabularx}
\end{table}

\subsection{Computational Resource Analysis}
\label{sec:computational}
To evaluate the computational cost of the model, we conducted a comparative analysis in both detection and anticipation tasks. 
For the VAD task, Uni-DSM operates directly on raw video clips without additional feature extractors. It uses a generative model only for score prediction rather than high fidelity generation, simplifying denoising and reducing computational cost. The runtime is mainly determined by the size of the noise conditional score transformer. In this setup, Uni-DSM uses an NCST with 130 million parameters and processes 8 frames in under 20 milliseconds, achieving 50 FPS. With the added computation of ByteTrack, the speed remains around 35 FPS, meeting typical real time surveillance requirements of about 30 FPS. We also compare inference speed and micro score with five representative methods including HF\(^2\)-VAD~\cite{liu2021hybrid}, VABD~\cite{li2021variational}, LLSH~\cite{lu2022learnable} , FPDM~\cite{yan2023feature} , and SSAE~\cite{cao2024scene} on the ShanghaiTech dataset. As shown in Table~\ref{tab:shanghaitech_resource}, Uni-DSM achieves the best performance of 84.5\% with competitive speed. Although VABD achieves the highest speed, it relies on pre-extracted optical flow, and its runtime would drop significantly if the extraction cost is considered. Our method captures characteristics of video anomalies while maintaining stable speed without additional input modalities.

For the VAA task, Uni-DSM demonstrates significant advantages in computational efficiency and resource utilization, as detailed in Table~\ref{tab:nwpu_resource}. Specifically, its GPU memory consumption is only 16 GB, substantially lower than the 21 GB of FBAE~\cite{cao2023new} and the 24 GB of SSAE~\cite{cao2024scene}, reflecting effective control over parameter count and computational complexity. Simultaneously, Uni-DSM achieves the fastest inference speed of 0.0163 seconds per frame, corresponding to 61.3 FPS, which markedly outperforms competing methods.  This indicates that Uni-DSM achieves both strong anticipation performance and high computational efficiency, making it suitable for real-world deployment under stringent real-time constraints.

\begin{figure}[t]
  \centering
  \includegraphics[width=0.98\columnwidth]{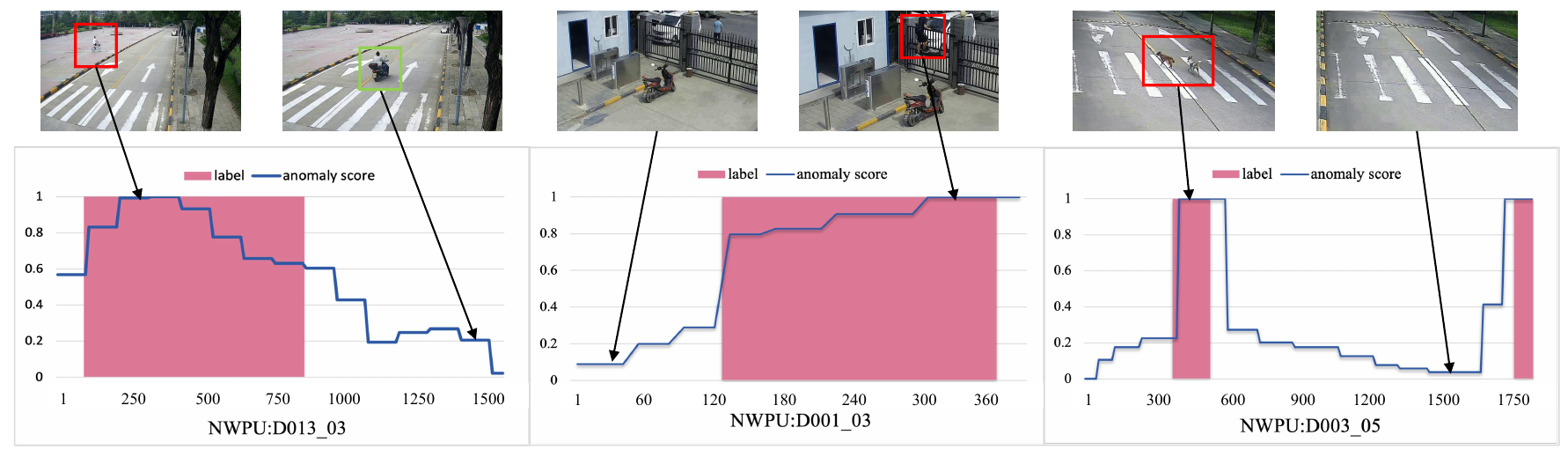}
  \caption{Visualization results on the NWPU dataset for the VAD task. The red boxes indicate that an anomaly is occurring, while the green boxes denote that the anomaly has disappeared.}
  \label{fig:fig10}
\end{figure}
\begin{figure}[t]
  \centering
  \includegraphics[width=0.98\columnwidth]{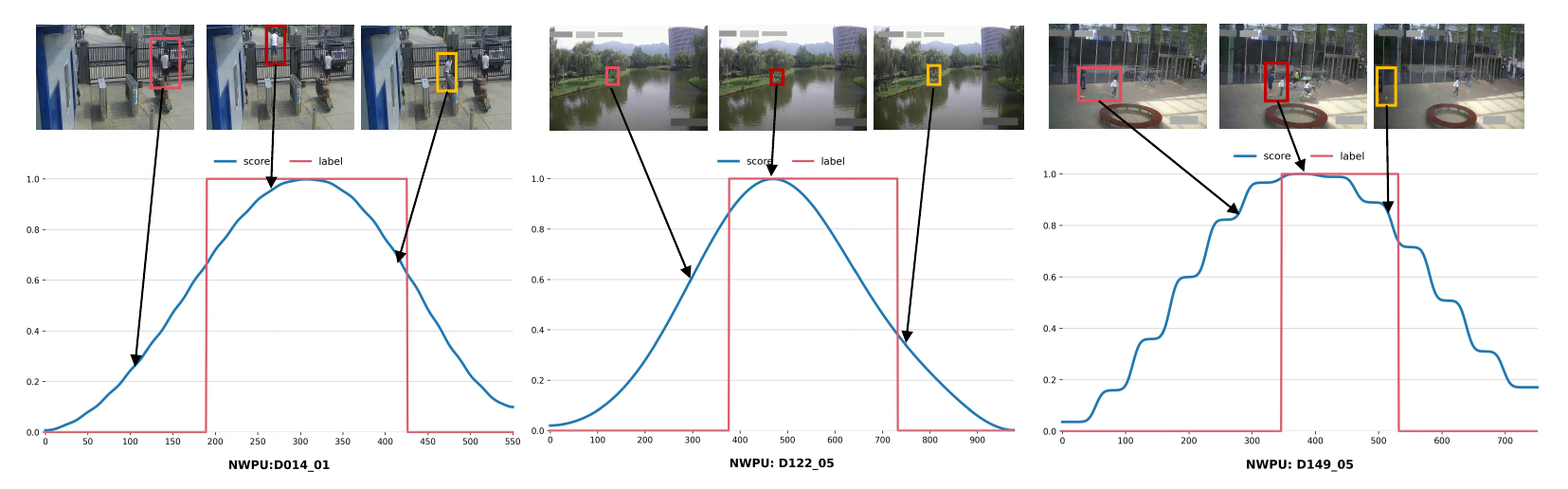}
  \caption{Visualization results on the NWPU dataset for the VAA task. The orange boxes indicate that an anomaly is about to occur, the red boxes denote that an anomaly is currently happening, and the yellow boxes signify that the anomaly is disappearing.}
  \label{fig:fig11}
\end{figure}

\subsection{Qualitative Analysis}
\label{sec:qualitative}
To provide intuitive insights into the detection and anticipation capabilities of Uni-DSM, we present qualitative visualizations of anomaly score dynamics. 
For the VAD task, we visualize anomaly score curves of ADSM on the NWPU dataset~\cite{cao2023new} for qualitative analysis. We select three clips to cover scene-dependent cyclist in D013\_03, motion anomaly climber in D001\_03, and appearance anomaly dog in D003\_05. As shown in Fig.~\ref{fig:fig10}, the cyclist that is abnormal on the plaza but normal on roads is detected promptly without false alarms. For the anomalous motion of the climber and the anomalous appearance of the dog, anomaly scores exhibit a sharp increase upon the occurrence of abnormal events and return to flat when the abnormal events disappear. These results show that the method captures the characteristics of video anomalies in terms of scene, motion, and appearance.

For the VAA task, we visualize anticipated anomaly score curves of SDSM for qualitative evaluation on three representative clips including wall climbing in D014\_01, lake disturbance in D122\_05, and theft in D149\_05. As shown in Fig.~\ref{fig:fig11}, the curves characterize the evolution of predicted anomaly likelihood within a given anticipation horizon. Notably, the anticipated curves are smoother than VAD curves, as SDSM performs anticipation over a continuous temporal window rather than discrete frames, resulting in reduced fluctuations. The score initially remains at a low baseline, then rises rapidly before anomaly onset and peaks around that point, stays elevated during the predicted interval, and gradually decreases as the scene returns to normal. 
This result highlights SDSM’s capability for proactive anomaly anticipation. Rather than passively reacting to observed events, the score evolves coherently with the anticipated emergence, progression, and resolution of anomalies. The early rise before visible onset reflects sensitivity to subtle precursory cues, while the sustained high values indicate continuous probabilistic forecasting over the anomaly duration. Meanwhile, consistently low scores on normal segments demonstrate effective suppression of false positives, and sharp transitions further reveal responsiveness to abnormal patterns.

\section{Conclusion}
In this paper, we present Uni-DSM, a unified score-driven framework grounded in denoising score matching that jointly addresses video anomaly detection and anticipation from a distributional perspective. Central to our approach is a shared noise-conditioned score transformer (NCST) backbone, enhanced by scene-dependent embeddings and motion-aware weights to ensure temporal consistency and robustness. For detection, we instantiate an autoregressive DSM (ADSM) mechanism that progressively accumulates anomalous context and integrates discrepancies between denoised outputs and original inputs to suppress local modes and enhance appearance perception. For anticipation, we propose a self-distilled DSM (SDSM) mechanism that employs a lightweight main-auxiliary architecture for unidirectional inference, constructing supervision from output discrepancies without relying on future ground truth. Extensive experiments demonstrate state-of-the-art performance across both tasks with improved inference speed and reduced computational overhead. Ultimately, our work establishes a scalable, unified paradigm that effectively integrates scene, motion, and appearance cues into a comprehensive score-driven framework.

\bibliographystyle{IEEEtran} 
\bibliography{main}
\begin{IEEEbiography}[{\includegraphics[width=1in,height=1.25in,clip,keepaspectratio]{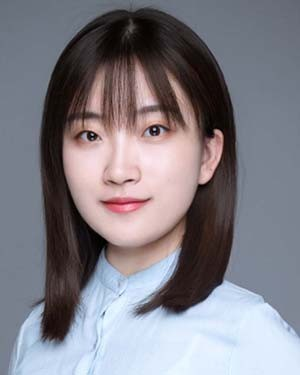}}]{Congqi Cao}
(Member, IEEE) received the BE degree in information and communication from Zhejiang University, Hangzhou, China, in 2013, and the PhD degree in pattern recognition and intelligent systems from the Institute of Automation, Chinese Academy of Sciences, in 2018. Then, she joined the School of Computer Science, Northwestern Polytechnical University, Xi'an, China, where she is currently an associate professor. Her research interests include computer vision, pattern recognition, and relative applications, especially action recognition/anticipation, video anomaly detection/anticipation, and few-shot learning.
\end{IEEEbiography}
\begin{IEEEbiography}[{\includegraphics[width=1in,height=1.25in,clip,keepaspectratio]{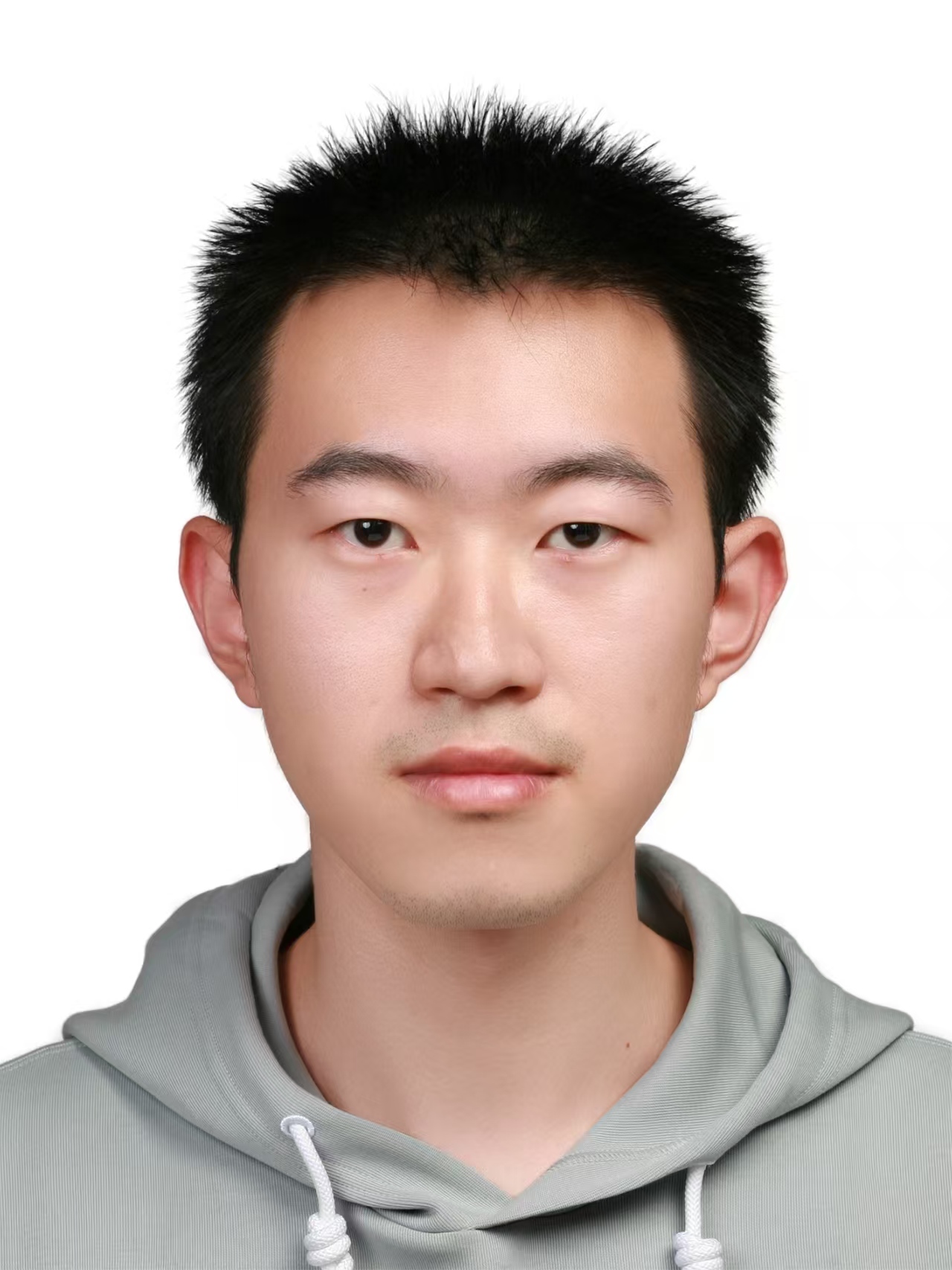}}]{Zhenhe Liang}
received the BE degree in data science and big data technology from Shandong University, Jinan, China, in 2024. He is currently working toward the ME degree in computer technology with the National Engineering Laboratory for Integrated Aero-Space-Ground-Ocean Big Data Application Technology, School of Computer Science, Northwestern Polytechnical University, Xi'an, China. His current research interests include computer vision and deep learning, especially video anomaly detection/anticipation and weather forecasting.
\end{IEEEbiography}
\begin{IEEEbiography}[{\includegraphics[width=1in,height=1.25in,clip,keepaspectratio]{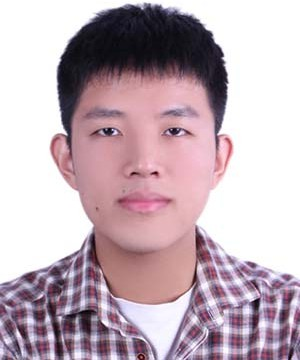}}]{Hanwen Zhang}
received the BE degree in computer science and technology from Northwestern Polytechnical University, Xi'an, China, in 2023, and the MS degree in computer science and technology from the School of Computer Science, Northwestern Polytechnical University, Xi'an, China, in 2026. His current research interests include computer vision and deep learning, especially video anomaly detection/anticipation.
\end{IEEEbiography}
\begin{IEEEbiography}[{\includegraphics[width=1in,height=1.25in,clip,keepaspectratio]{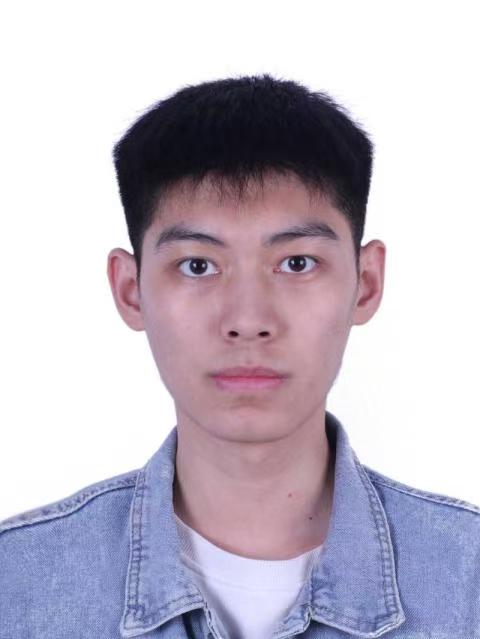}}]{Yifan Zhao}
received the BE degree in computer science and technology from Northwestern Polytechnical University (NWPU), Xi'an, China, in 2025. He is currently working toward the ME degree in computer technology with the National Engineering Laboratory for Integrated Aero-Space-Ground-Ocean Big Data Application Technology, School of Computer Science, Northwestern Polytechnical University, Xi'an, China. His main research interests include video anomaly detection.
\end{IEEEbiography}
\begin{IEEEbiography}[{\includegraphics[width=1in,height=1.25in,clip,keepaspectratio]{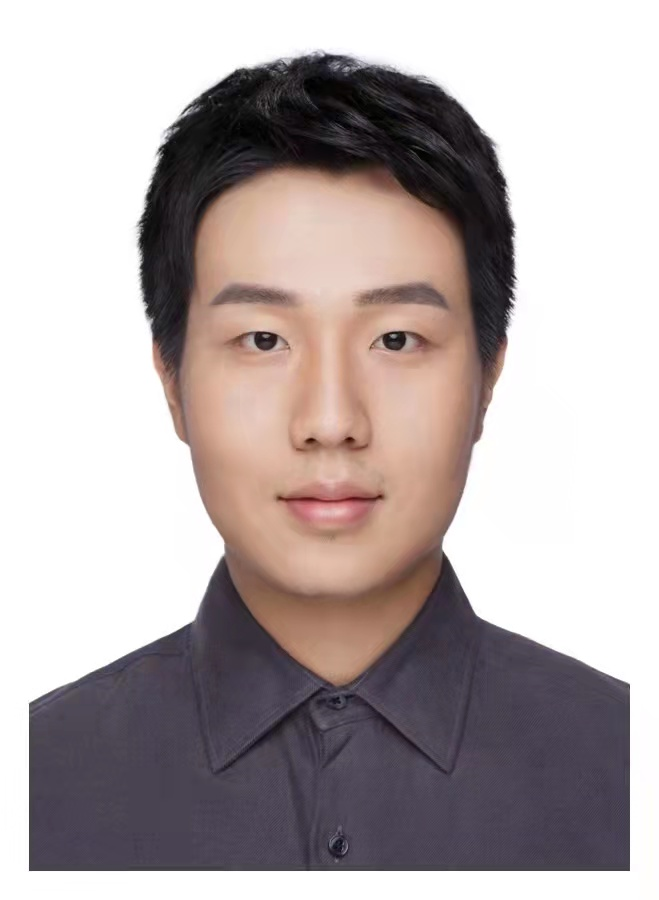}}]{Qinyi Lv}
received the BS and PhD degrees from Zhejiang University, Hangzhou, China, in 2013 and 2018, respectively. He is an associate professor with Northwestern Polytechnical University. His recent research interests include intelligent sensing, intelligent systems, radar detection, and inverse scattering problems.
\end{IEEEbiography}

\begin{IEEEbiography}[{\includegraphics[width=1in,height=1.25in,clip,keepaspectratio]{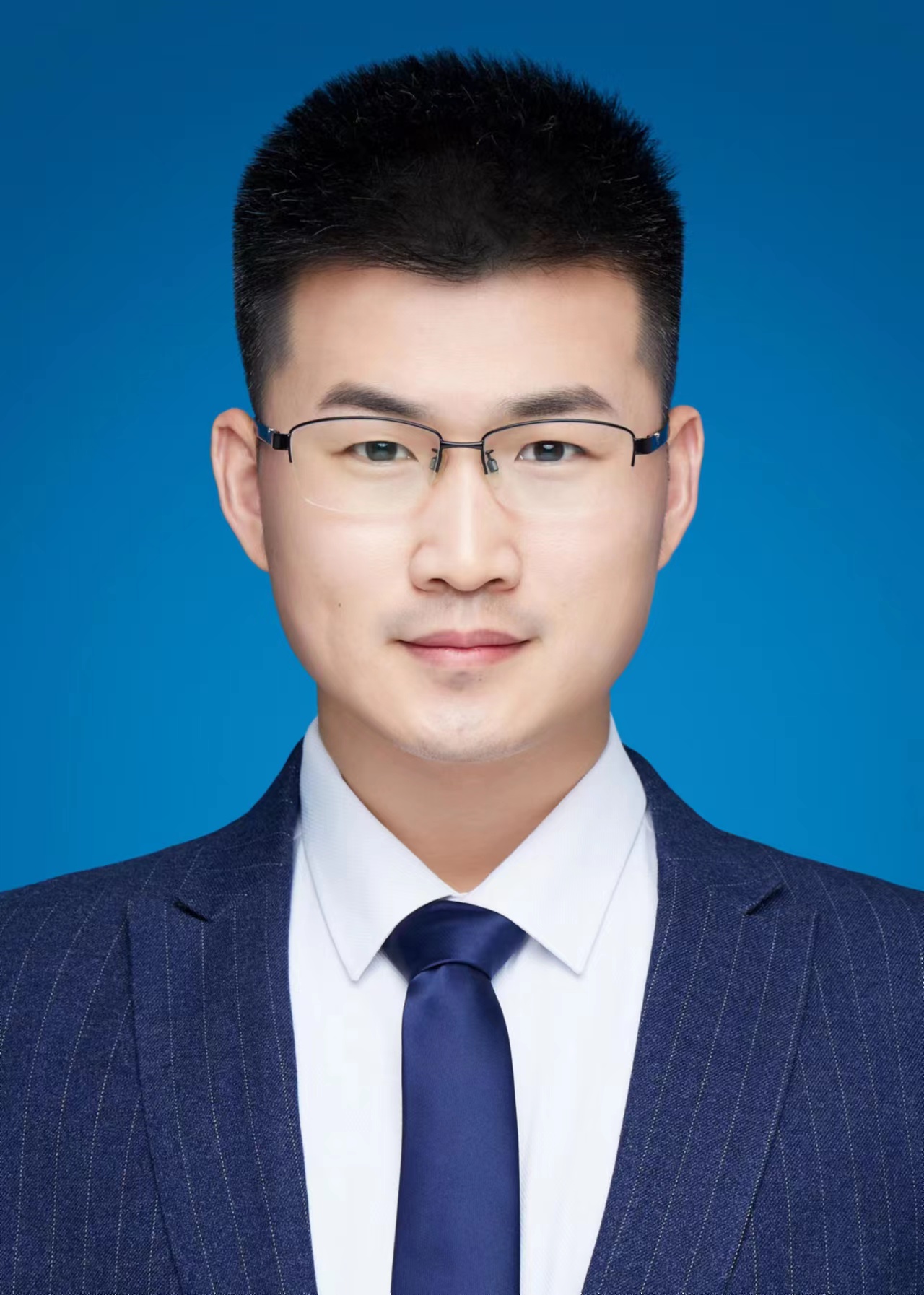}}]{Lingtong Min}
(Member, IEEE) received the BS degree from Northeastern University, Shenyang, China, in 2012, and the PhD degree from Zhejiang University, Hangzhou, China in 2019. He is a professor with Northwestern Polytechnical University. His main research interests are computer vision, pattern recognition, and remote sensing image understanding.
\end{IEEEbiography}
\newpage
\begin{IEEEbiography}[{\includegraphics[width=1in,height=1.25in,clip,keepaspectratio]{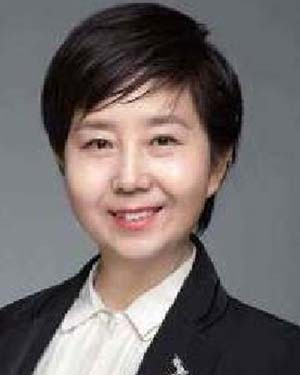}}]{Yanning Zhang}
(Fellow, IEEE) received the B.S. degree from Dalian University of Technology, Dalian, China, in 1988, the M.S. degree from the School of Electronic Engineering, Northwestern Polytechnical University, Xi’an, China, in 1993, and the Ph.D. degree from the School of Marine Engineering, Northwestern Polytechnical University, in 1996. She is currently a Professor with the School of Computer Science, Northwestern Polytechnical University. She is also a Cheung Kong Professor with the Ministry of Education, China. She has authored more than 200 articles. Her current research interests include remote sensing image analysis, computer vision, and pattern recognition. She is an Associate Editor of IEEE Transactions on Geoscience and Remote Sensing.
\end{IEEEbiography}
\end{document}